\documentclass[twoside,11pt]{article}

\usepackage[accepted]{melba}

\usepackage{amsmath,amsfonts}
\usepackage{latexsym}
\usepackage{amsmath}
\usepackage{mathtools}
\usepackage{algorithmic}
\usepackage{algorithm}
\usepackage{textcomp}
\usepackage{lipsum}
\usepackage{xcolor}
\usepackage{caption}
\usepackage{subcaption} 
\usepackage{lipsum}
\usepackage{nicefrac}
\usepackage{enumitem}

\newcommand{\ie}{\emph{i.e.}~}
\newcommand{\cf}{\emph{cf.}~}

\newcommand{\tariq}[1]{\textcolor{black}{#1}}

\usepackage{pgfplots}
\usepackage{tikz}
\definecolor{bblue}{HTML}{4F81BD}
\definecolor{rred}{HTML}{C0504D}
\definecolor{ggreen}{HTML}{9BBB59}
\definecolor{ppurple}{HTML}{9F4C7C}

\definecolor{new1}{HTML}{9A4C7B}

\definecolor{new2}{HTML}{AFC47C}

\melbaheading{2022:011}{https://www.melba-journal.org/papers/2022:011.html}{2022}{1-37}{11/2021}{03/2022}{Bdair, Navab, and Albarqouni}{}{}

\ShortHeadings{FedPerl for Skin Lesion Classification}{Bdair, Navab, and Albarqouni}
\firstpageno{1}

\title{Semi-Supervised Federated Peer Learning for Skin Lesion Classification}

\author{\name Tariq M. Bdair \email t.bdair@tum.de \\  
	\addr Computer Aided Medical Procedures, Technical University of Munich, Germany
	\AND
	\name Nassir Navab\email nassir.navab@tum.de \\
	\addr Computer Aided Medical Procedures, Technical University of Munich, Germany \\ 
	The Whiting School of Engineering, Johns Hopkins University, United States 
	\AND
	\name Shadi Albarqouni\email shadi.albarqouni@ukbonn.de \\
	\addr Clinic for Diagnostic and Interventional Radiology, University Hospital Bonn, Germany\\
	Helmholtz AI, Helmholtz Zentrum München, Germany \\
	Faculty of Informatics, Technical University of Munich, Germany \\ 
}

\begin{document}

\maketitle

\begin{abstract}
Globally, Skin carcinoma is among the most lethal diseases. Millions of people are diagnosed with this cancer every year. Sill, early detection can decrease the medication cost and mortality rate substantially. The recent improvement in automated cancer classification using deep learning methods has reached a human-level performance requiring a large amount of annotated data assembled in one location, yet, finding such conditions usually is not feasible. Recently, federated learning (FL) has been proposed to train decentralized models in a privacy-preserved fashion depending on labeled data at the client-side, which is usually not available and costly. To address this, we propose \verb!FedPerl!, a semi-supervised federated learning method.  Our method is inspired by peer learning from educational psychology and ensemble averaging from committee machines. \verb!FedPerl! builds communities based on clients' similarities. Then it encourages communities' members to learn from each other to generate more accurate pseudo labels for the unlabeled data. We also proposed the peer anonymization (PA) technique to \tariq{anonymize clients}. As a core component of our method, PA is orthogonal to other methods without additional complexity, and reduces the communication cost while enhances performance. Finally, we propose a dynamic peer learning policy that controls the learning stream to avoid any degradation in the performance, especially for the individual clients. Our experimental setup consists of 71,000 skin lesion images collected from 5 publicly available datasets. {We test our method in four different scenarios in \verb!SSFL!. With few annotated data, \verb!FedPerl! is on par with a state-of-the-art method in skin lesion classification in the standard setup while outperforming \verb!SSFLs! and the baselines by 1.8\% and 15.8\%, respectively.} Also, it generalizes better to an unseen client while being less sensitive to noisy ones.
\end{abstract}


\section{Introduction}
In the recent estimation, an expectation of two hundred thousand fatal invasive and in-situ melanoma cases will be diagnosed in the USA in 2021~\citep{tarver2021cancer}. Yearly, millions of people are diagnosed with skin carcinoma ~\citep{rogers2010incidence}.  Worldwide, skin cancer is considered one of the most expensive and fatal cancers.
While most non-melanoma skin cancer cases can be cured, the melanoma ones are curable when detected in the early stages. For example, the 5-year survival rate ranges from 99$\%$ in the earliest stage to 27$\%$ for the latest stage~\citep{tarver2021cancer}. Moreover, early detection of skin cancer can reduce the treatment expenses significantly~\citep{esteva2017dermatologist}. Therefore, several attempts have investigated the automated classification of skin lesions in dermoscopic images~\citep{clark1989model,binder1998epiluminescence,schindewolf1993classification}. Though, these attempts require handcrafted engineered features and exhausted pre-processing steps.

Yet, huge improvements in computerized methods have been achieved in recent years. For instance, the deep-learning-based methods proved to have a superior \citep{gessert2020skin,li2020domain,zhang2019attention,lopez2017skin} or a human-level performance \citep{esteva2017dermatologist,tschandl2019comparison} when dealing with skin cancer classification.
Nevertheless, this success comes at the cost of exhausting pre-processing steps,  a prudently designed framework, or a substantial amount of labeled data assembled in one location. In real life, medical data is generated from different scanners and unevenly distributed in multiple centers in raw formats without annotations resulting in heterogeneous data, or so-called Non-IIDness. Unfortunately, building a large repository of annotated medical data is quite challenging due to privacy burdens~\citep{rieke2020future,kaissis2020secure}, and labeling cost which is time-consuming and requires domain expert knowledge.

Federated learning (FL) \citep{mcmahan2017communication} has been recently proposed to learn machine learning models utilizing the ample amounts of labeled data distributed in mobile devices while maintaining clients' privacy \ie without sharing the data. 
The training process of federated learning starts at the server by broadcasting initial weights of global model parameters to a random set of participating clients, who share the same model architecture with the global model. Each client, afterward, trains locally on its local data before sending back the updated model parameters to the server. Once all clients send their updates, the server aggregates them using \verb!FedAvg! to update the global model weights. Next, the updated global model is broadcasted to a new random set of clients before a new round of local training processes starts. Eventually, the previous steps are repeated until the global model converged. During the training, only model weights are shared while data is kept locally. Note that, the key properties of the FL are data privacy, Non-IIDness, and communication efficiency. Thus, FL goes in line with the nature of the medical setting.
Consequently, federated learning has been investigated by several works in the medical domain \citep{zhu2019privacy,li2020federated,albarqouni2020domain} paving the way to training machine learning models in privacy-preserved fashion in real-world applications~\citep{flores2021federated,roth2020federated,sarma2021federated}. Though, in the previous works, the training demand highly accurate labeled data, e.g., ground-truth confirmed through histopathology, which often is costly and not available. 

In a more realistic scenario, the clients may have access to a large amount of unlabeled data along with few annotated ones. Yet, willing to train a reliable model to make use of their data. Fortunately, the above scenario can be addressed by the semi-supervised learning (SSL) paradigm, which is the focus of this paper. In this regard, a very recent work \citep{yang2021federated} has shown the applicability of semi-supervised learning in a federated setting (a.k.a. \verb!SSFL!) for COVID-19 pathology segmentation. The previous work among the firsts who introduced semi-supervised learning to federated learning. Yet, they have straightforwardly applied a semi-supervised learning method, \emph{e.g.} \verb!FixMatch! \citep{sohn2020fixmatch} locally. At first, a local model is trained in a fully supervised fashion using the labeled data.  Then, the trained model is used to produce predictions for unlabeled data, where the predictions with high confidence are used to generate pseudo labels. Next, the pseudo labels are attached to the labeled data before a new training process starts. At the server, on the other hand, \verb!FedAvg! was employed to organize the training between different clients, see Sec.~\ref{SSFL}. {Another recent work, \citep{liu2021federated} proposed an \verb!SSFL! approach; \verb!FedIRM! for skin lesion classification. They suggested distilling the knowledge from labeled clients to unlabeled ones through building a disease relation matrix, extracted from the labeled clients, and providing it to the unlabeled ones to guide the pseudo-labeling process.} In a more challenging situation, which has not been yet investigated thoroughly in the medical images, the labeled data is located at the server side while the clients have access only to unlabeled data. This scenario has been addressed in this paper, see Sec.~\ref{secUClient}.

In \verb!SSFL!, clients are only trained i) globally, where the knowledge is accumulated in global model parameters, and ii) locally, where the knowledge is distilled via the local data. While this is a simple and straightforward approach, we argue that the knowledge gain for generating pseudo labels for the local models is limited. Instead, we hypothesize that gaining extra knowledge by learning from similar clients \ie Peer Learning (PL) is highly significant assuming that peer learning encourages the self-confidence of the clients by sharing their knowledge in a way that does not expose their identities.   

Our method is highly inspired by the social science literature, where peer learning is defined as acquiring skills and knowledge through active helping among the companions. It involves people from similar social groups helping each other to learn \citep{topping2005trends}. Peer Learning includes peer tutoring, coaching, mentoring, and others. Though, the distinguishing between these types of PL is out of the scope of this paper. In this work, the link between peer learning and federated learning is direct, where the clients are considered as peers learning from each other.

In the computer science literature, a similar concept to peer learning has been introduced known as Committee Machines (CM)~\citep{tresp2001committee,aubin2019committee,joksas2020committee}. In a nutshell, CM is a well-known and active research direction and is defined as an ensemble of estimators, consisting of neural networks or committee members, that cooperate to obtain better accuracy than the individual networks. The committee prediction is generated by ensemble averaging (EA) of the individual members' predictions. CM has shown to be effective in machine learning hardware \citep{joksas2020committee}, yet, it has not been investigated in FL.
In this work, we employ the ensemble averaging from the committee machine for our \textit{peers anonymization} (PA) technique. PA improves privacy by hiding clients' identities. Moreover, PA reduces the communication cost while preserving performance. To the best of our knowledge, no prior work has proposed the PA technique in the \verb!SSFL! for the medical images.

Peer learning has shown an increase in the performance for the majority of clients, see Sec.~\ref{exp:CLR}. However, for some clients, peer learning and federated models could harm their performance. Thus, we propose a dynamic peer-learning policy that controls the learning process. Our dynamic learning policy maintains the performance of all clients while boosting the accuracy of the individual ones, see Sec.~\ref{secLP}.

In this paper, we propose \verb!FedPerl!, where the key properties are peer learning, peer anonymization, and learning policies. Our approach, in contrast to a very recent work FedMatch~\citep{jeong2020federated}, is communication-efficient and \tariq{hides clients' identities} where it employs the ensemble averaging method before sharing clients' knowledge with other peers. 
Our first contributions that include peer learning and peer anonymization in the standard semi-supervised learning setting were presented in ~\citep{bdair2021fedperl}. In this extended and comprehensive version, we add the following \textbf{contributions}:
\begin{itemize}
\item We propose a dynamic learning policy that controls the contribution of peer learning in the training process. While our dynamic policy excels the static one, it at the same time helps the individual clients to achieve better performance.
\item We show that our peer anonymization is orthogonal and can be easily integrated into other methods without additional complexity. 
\item We introduce and test our method in a challenging scenario, not been yet investigated thoroughly in the medical images, where the labeled data is located at the server-side. Moreover, we test the ability of FedPerl to generalize to unseen clients. Additionally, we conduct extensive analyses on the effect of committee size on the performance at the client and community levels. 
\item We introduce additional evaluation metrics to evaluate the calibration of these models and their clinical applicability. 
\item We validate our method on skin lesion classification, with database consists of more than 71,000 images, showing superior performance over the baselines.
\end{itemize}

\section{Methodology}
\begin{figure*}[!t]
\centering
\includegraphics[width=1\textwidth]{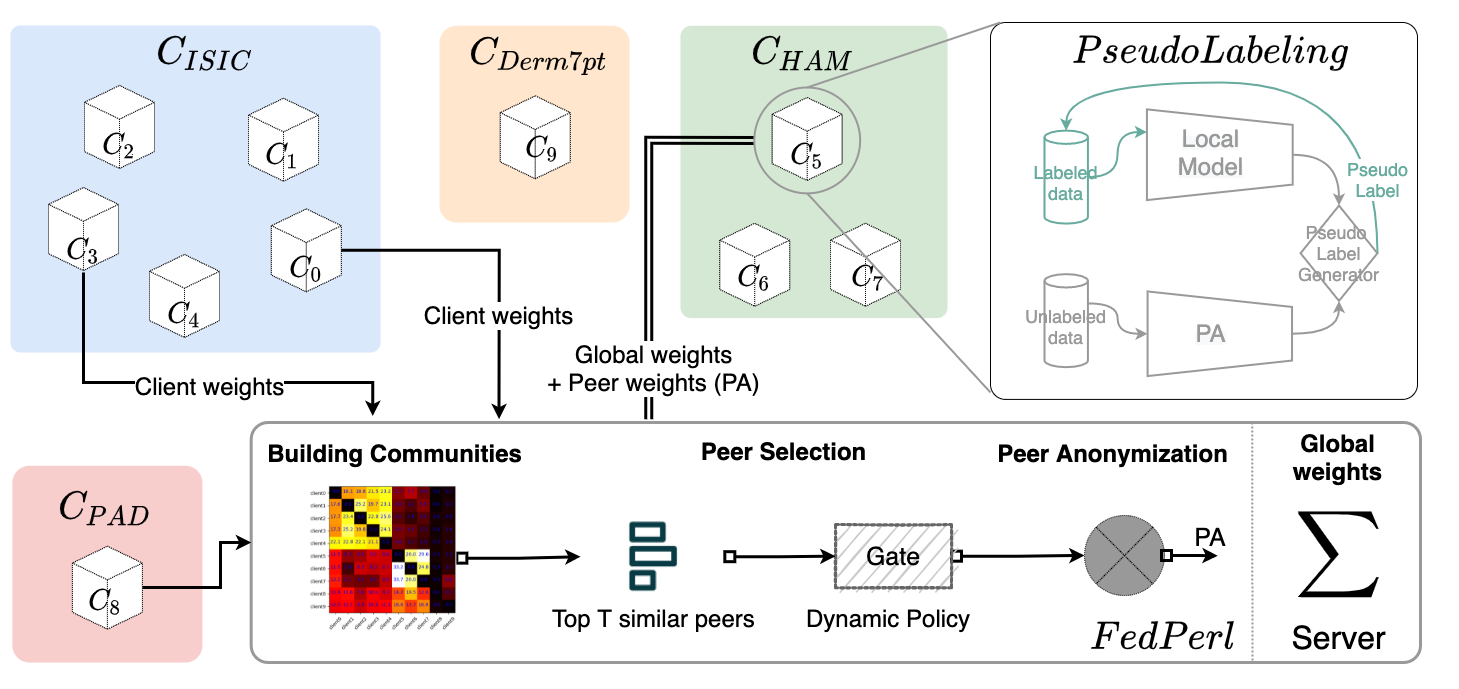}
\caption{Our semi-supervised federated learning framework (FedPerl). Our method consists of (i) Building communities: similar clients clustered into one community, (ii) Peer Learning: peers are helping in pseudo labeling, and (iii) Peer Anonymization (PA) to hide client identity, improve privacy, and reduce the communication cost. Top Right: Pseudo labeling utilizing an anonymized peer in this diagram. Bottom: Selecting the similar peers, peer learning, peers anonymization, and similarity matrix calculations are performed on the server. FedPerl exploits either static or dynamic learning policies.}
\label{figframe}
\end{figure*}
\subsection{Problem Formulation}
\label{method:PF}
Given $M$ clients $\mathcal{C}_{m}$ who have access to their own local dataset $\mathcal{D}_{m} \in \mathbb{R}^{H\times W\times N_{m}}$, where $H$ and $W$ are the height and the width of the input images, and $N_{m}$ is the total number of images.  
$\mathcal{D}_{m}$ consists of labeled $\mathcal{S}_L = \{ \mathcal{X}_{L},  \mathcal{Y}_{L} \}$ and unlabeled data $\mathcal{S}_{U} = \{ \mathcal{X}_{U}\}$, where $\{ \mathcal{X}_{L},  \mathcal{X}_{U} \} = \{\textit{\textbf{x}}_1,\ldots, \textit{\textbf{x}}_L, \textit{\textbf{x}}_{L+1},\ldots, \textit{\textbf{x}}_{L+U}\}$ are input images; $\textit{\textbf{x}}\in \mathbb{R}^{H\times W}$, and $\mathcal{Y}_{L} = \{\textit{\textbf{y}}_1,\ldots,\textit{\textbf{y}}_L\}$; $\textit{\textbf{y}}\in \mathbb{R}^{C}$ are the corresponding categorical labels for $C$ classes. Given query image \textit{$\textbf{x}_q$}, our objective is to train a global model $f(\cdot)$ to predict the corresponding label \textit{$\tilde{\textbf{y}_q}$} for \textit{$\textbf{x}_q$}, where labeled and unlabeled data are leveraged in the training in a privacy-preserved fashion.
\\

\textbf{Definition.}
\label{sec:def}
We define a model $f(\cdot)$ to be trained in a privacy-preserved fashion, if the following conditions are met \\

\textit{
(i) Data can not be transferred across different clients participating in the training process adhering to the General Data Protection Regulation (GDPR)~\footnote{\url{https://gdpr.eu/}}.} \\

\textit{
(ii) Local models can not be transferred across different clients participating in the training process to avoid privacy breaches~\citep{orekondy2018gradient}, or model inversion~\citep{fredrikson2015model}.}

\subsection{Semi-Supervised Federated Learning (SSFL)}
\label{SSFL}
The aforementioned conditions can be met by picking off-the-shelf SoTA SSL models, \emph{e.g.,} \verb!FixMatch!~\citep{sohn2020fixmatch}, to train the clients locally leveraging the unlabeled data, while employing \verb!FedAvg!~\citep{mcmahan2017communication} to coordinate between the clients in a federated fashion as \citep{yang2021federated},
\begin{equation}
\min_{\phi}\mathcal{L}(\mathcal{D}; \phi) \quad \text{with} \quad  \mathcal{L}(\mathcal{D}; \phi)=\sum_{m=1}^{M}w_m\;\mathcal{L}_{{SSL}_{m}}(\mathcal{D}_{m}; \phi),
\label{eq:formalism}
\end{equation}
where $w_m=\nicefrac{N_m}{\sum_{i=1}^{M} N_i}$ is the respective weight coefficient for each client, and $\phi$ is the model parameters. The SSL objective function appeared in \verb!FixMatch!~\citep{sohn2020fixmatch}, can be used to train the client locally utilizing both labeled and unlabeled data as
\begin{equation} \label{eq1}
\begin{split}
\mathcal{L}_{{SSL}_{m}}(\mathcal{D}_{m}; \phi) = \arg\min_{\phi} \mathcal{L}_{CE}(\mathcal{Y}_{L}, f(\alpha(\mathcal{X}_{L});\phi)) \\ +  \beta\mathcal{L}_{CE}(\mathcal{\Tilde{Y}}_{U},f(\mathcal{A}(\mathcal{X}_{U});\phi)) , 
\end{split}
\end{equation}
where $\mathcal{L}_{CE}(\cdot,\cdot)$ is the cross-entropy loss, $\beta$ is a hyper-parameter that controls the contribution of the unlabeled loss to the total loss, $\mathcal{\Tilde{Y}}_{U}$ is the pseudo labels for the unlabeled data $\mathcal{X}_{U}$, and $\alpha(\cdot)$ and $\mathcal{A}(\cdot)$ are weak and strong augmentations respectively. 
For an unlabeled input $\textit{\textbf{x}}_{i}$, the pseudo label $\tilde{\textit{\textbf{y}}}_{i} \in \mathcal{\tilde{Y}}_{U}$, is produced by applying a confidence threshold $\tau$ on the client' prediction on a  weak augmented version of $\textit{\textbf{x}}_{i}$ such that
\begin{equation}
\label{eq:PL}
\tilde{\textit{\textbf{y}}}_{i} = \arg \max (\mathbb{I} (f_{\mathcal{C}_{m}}(\alpha(\textit{\textbf{x}}_{i});\phi^{*}) \geq \tau)),
\end{equation}
where $f_{\mathcal{C}_{m}}(.)$ is the local model, $\phi^{*}$ are frozen model parameters, and $\mathbb{I}(\cdot)$ is the indicator function.
\subsection{FedPerl: Peer Learning in SSFL}
\label{method:FedPer}
While the straightforward \verb!SSFL! is simple, we argue that the learned knowledge for the individual clients could be further improved by involving similar clients in the training.  Inspired by peer learning, our method utilizes similar peers to help the target client in the pseudo labeling by sharing their knowledge without \tariq{exposing their identities} through employing the peer anonymization method. 
Our proposed \verb!FedPerl!, illustrated in Fig.~\ref{figframe}, consists of three components; namely 1) building communities, 2) peer learning, and 3) peer anonymization. While peer learning can be static or dynamic as shown in the following sections. 
\subsubsection{Building communities}
\label{method:BC}
In educational social science~\citep{topping2005trends}, "peers" are referred to as two or more persons who share similarities and consider themselves as companions. In this work, we adopt the same concept and describe a group of clients as "peers" if they are similar. 
Previous work has shown that clustering can be achieved using models updates \citep{briggs2020federated}, While other works  measure similarities between deep neural networks by comparing the representations between layers \citep{kornblith2019similarity}. We build upon this and argue that the model weights represent and summarize the learned knowledge for each client from its training data.
Thus, to measure the similarities between the clients, we represent each client $\mathcal{C}_{m}$ by a feature vector $\mathcal{F}_{m}= \{(\mu_{0}, \sigma_{0}), \ldots,(\mu_{l}, \sigma_{l})\} \in \mathbb{R}^{2\cdot l}$, where $(\mu_{l}, \sigma_{l})$ is the first two statistical moments, \ie the mean and the standard deviation, of the model's layer $l$ parameters. Then, we compute the similarity $\omega_{mk}$ between clients $\mathcal{C}_{m}$ and $\mathcal{C}_{k}$ using the cosine similarity, where $\omega_{mk} = \frac {\mathcal{F}_{m}^T\mathcal{F}_{k}}{\|\mathcal{F}_{m}\| \cdot \|\mathcal{F}_{k}\|}$. Using the cosine similarity brings the model parameters to the same behaviors without being exact, given that the means might differ, as long as they are in the same direction. Finally, the similarity matrix between all clients is defined as
\begin{align} \label{eq2}
\mathcal{W}_{M\times M} = \begin{bmatrix}
    \omega_{11} & \ldots &\omega_{1M} \\
    \vdots & \ddots & \vdots \\
    \omega_{M1} & \ldots &\omega_{MM}
\end{bmatrix}.
\end{align}

Our method starts with standard federated learning warm-up rounds (\emph{e.g.} ten rounds in our case). In the next training rounds, the feature vectors are extracted after receiving the updates from the participating clients. Then, the similarity matrix is computed and updated accordingly. In \verb!FedPerl!, The communities are formed implicitly based on the similarity matrix where similar clients are clustered into one community (see Sec.~\ref{exp:BC}).
\subsubsection{Peers Learning}
The term "learning" is frequently defined as improved knowledge, experiences, and capabilities \citep{topping2005trends}. 
In peer learning, "peers" help each other by sharing their knowledge \citep{topping2005trends}. 
In this regard, we describe "peer learning" as the means of top $T$ alike clients (peers) help each other to generate pseudo labels by sharing their knowledge (model parameters).
This is a helpful process since a main property of the medical data is the data heterogeneity. 
In federated learning, the clients experience different data and class distribution during the training. Thus, accumulating and sharing the distributed knowledge is useful. Particularly, it can help the local client generate pseudo labels for the unlabeled data from experiences that might never have learned from its own labeled data. To realize this, we modify the pseudo label defined in Eq.\ref{eq:PL} to include the predictions of the similar $T$ peers, \emph{i.e.} $f_{t}(\cdot;\phi)$ according to the similarity matrix $\mathcal{W}$ as
\begin{align} \label{eq3}
\tilde{\textit{\textbf{y}}}_{i} = \arg \max \left(\mathbb{I} \left(f_{\mathcal{C}_{m}}(\alpha(\textit{\textbf{x}}_{i});\phi^*) + \sum_{t=0}^{T} f_{t}(\alpha(\textit{\textbf{x}}_{i});\phi_t^*) \geq \tau\right)\right).
\end{align}
\subsubsection{Peers Anonymization} 
To improve privacy and adhering to the privacy regulations introduced in \ref{sec:def}, the knowledge sharing among peers has to be anonymized and regulated. Thus, we propose \textit{peers anonymization} (PA), at the server side, a simple, yet effective technique.  Particularly, we create an anonymized peer $f_{a}(\cdot; \phi_a)$ that assembles the learned knowledge from the top $T$ similar peers where 
\begin{align} \label{eq4}
f_{a}(\cdot;\phi_a) = \frac{1}{|T|}\sum_{t=0}^{T} f_{t}(\cdot; \phi_t).
\end{align}
Then, $f_{a}(\cdot)$ is shared with the local model to help in pseudo labeling. Accordingly, Eq.\ref{eq3} is modified to
\begin{align} \label{eq5}
\tilde{\textit{\textbf{y}}}_{i} = \arg \max \left(\mathbb{I} \left(f_{\mathcal{C}_{m}}(\alpha(\textit{\textbf{x}}_{i});\phi^*) +  f_{a}(\alpha(\textit{\textbf{x}}_{i});\phi_a^*) \geq \tau\right)\right).
\end{align}
Notice that sharing the peers and the anonymized peer are not equivalent (Sec.\ref{exp:PLR}), \emph{i.e.} $\frac{1}{|T|}\sum_{t=0}^{T} f_{t}(\alpha(\textit{\textbf{x}}_{i}); \phi_t)) \neq f_{a}(\alpha(\textit{\textbf{x}}_{i});\phi_a)$.
Eventually, the anonymized peer is shared only one time for each client at every training round, not at every local update. The advantages of the anonymized peer are i) it reduces the communication cost as sharing the knowledge of one peer is better than sharing 2 or more peers, ii) hides clients' identities by creating an anonymized peer.
Finally, to prevent the local model from deviated from its local knowledge, we employ an MSE loss as a consistency-regularization term, which broadly used in semi-supervised learning,
\begin{align} 
\label{eq6}
\mathcal{L}_{{CON}_{m}} = \parallel f_{\mathcal{C}_{m}}(\textit{\textbf{x}}_{i};\phi) -  f_{a}(\textit{\textbf{x}}_{i};\phi_a^*) \parallel^2.
\end{align}
\subsubsection{Overall objective:}
The overall objective function for client $m$ is the sum of semi-supervised and consistency-regularization losses, and given by
\begin{align} \label{eq7}
\mathcal{L}_{m} =  \mathcal{L}_{{SSL}_{m}} + \gamma \mathcal{L}_{{CON}_{m}},
\end{align}
where $\gamma$ is a hyperparameter, and $\mathcal{L}_{{SSL}_{m}}$ and $\mathcal{L}_{{CON}_{m}}$ are Eq.\ref{eq1} and Eq.\ref{eq6}, respectively. Note that the two terms in Eq.\ref{eq7} collaborate to achieve the balance between the local and global knowledge.

\subsubsection{Dynamic Learning Policy}
\label{method:DLP}
Thus far, we have proposed a static learning policy in which the top $T$ similar peers are used to help the local clients in the pseudo labeling process. In peer learning, the clients are divided into groups or communities based on their similarities. A natural result of this step is also individual clients who do not belong to any community. Practically, we may have no control over the effect of applying the static peer learning policy on these clients, which could vary from one client to another where it is beneficial for some clients and not for others. For example, individual clients who do not belong to any community would be forced to learn from top $T$ peers, based on the proposed similarity matrix, however, there is no guarantee that they would be beneficial in the training since they may not belong to the same or similar community. Therefore, we suggest performing a dynamic policy where we could carefully involve the peers based on additional similarities or restricting the peers to a subset who are close enough. Our dynamic learning policy controls the learning stream to the clients where the peers are utilized in the learning process. In short, our goal is to maintain the gain and boost the performance for all clients. In this regard, we propose the following policies. 
\paragraph{\noindent\textbf{Validation Policy}} In this policy, first, the client and its peers are validated on the global validation dataset. Then, only the peers with a validation accuracy equal to or higher than the client's accuracy are utilized. This policy can be applied with or without the peers' anonymization technique. Formally, assume that $V_{acc}(.)$ is a function that measures the accuracy on a global validation dataset, then the set of the peers that participate in peer learning for client $\mathcal{C}_{m}$ is defined as  
\begin{align} 
\label{eqValPol}
\Omega =  \{\mathcal{C}_{n} | V_{acc}(\mathcal{C}_{n}) \geq V_{acc}(\mathcal{C}_{m})\},
\end{align}
where $\mathcal{C}_{n}$ is a peer, $n=1,2,..., T$, and $T$ is the committee size. 
\paragraph{\noindent\textbf{Gated Validation Policy}} As in the previous policy, the peers are validated on the global validation dataset. However, we apply a gateway on their accuracies, such that if it is equal to or higher than a pre-defined gateway threshold $\rho$, the peer will be involved in the pseudo labeling. Otherwise, it will be discarded from the process. In this policy, the set of the peers that participate in peer learning for client $\mathcal{C}_{m}$ is defined as    
\begin{align} 
\label{eqGatVal}
\Omega =  \{\mathcal{C}_{n} | V_{acc}(\mathcal{C}_{n}) \geq \rho\}.
\end{align}
\paragraph{\noindent\textbf{Gated Similarity Policy}}Like in the gated validation policy, this policy depends on a gateway that controls peers participation. Yet, no validation set is used, and a peer is allowed to participate if its similarity with the client is equal to or higher than the gateway threshold $\rho$. Assume that $H_{sim}(.)$ is a function that measures the similarity between two clients, then the set of the peers that participate in peer learning for client $\mathcal{C}_{m}$ is defined as  
\begin{align} 
\label{eqValPol}
\Omega =  \{\mathcal{C}_{n} | H_{sim}(\mathcal{C}_{m},\mathcal{C}_{n}) \geq \rho)\}.
\end{align}
Note that regardless of the used policy, we first select the top $T$ similar peers based on the similarity matrix. Then, one of the above policies is applied. The only difference between the last two policies is that in the gated validation policy, we used the validation accuracy as a gateway, while in the gated similarity policy, we stick to our similarity matrix. A pseudo-code summarizing our method is shown in Algorithm \ref{alg:algo1} 
\begin{algorithm}[t]
\caption{Semi-Supervised Federated Peer Learning for Skin Lesion Classification}
\label{alg:algo1}
\begin{algorithmic}[1]
\STATE \textbf{StartServer()}

\STATE {initialize global weights $\Phi_{G}^{0}$}

\STATE {\textbf{for each} round r=1, 2, ..., $R$ \textbf{do:}}

\STATE {{\quad $n$ $\longleftarrow$ select random $n$ clients from $M$} \textcolor{gray}{// \ie n=3}}

\STATE {{\quad \textbf{for each} client $\mathcal{C}_{m}$ in $1, 2, ..., n$ \textbf{do in parallel:}}}

\STATE {{\quad \quad $f_{\mathcal{C}_{m}} \longleftarrow$ initialize client's weights with global weights}}

\STATE {{\quad \quad \textbf{if} $r > 10:$} \textcolor{gray}{//Peer learning starts after warm-up rounds}}

\STATE {{\quad \quad \quad $f_{\mathcal{C}_{1:T}} \longleftarrow $ GetTopSimilarPeers($T, f_{\mathcal{C}_{m}}$)}  \textcolor{gray}{// Sec. \ref{method:BC}}} 

\STATE {{\quad \quad \quad if $\textbf{IsDynamicPolicy}$} \textcolor{gray}{// Sec. \ref{method:DLP}}} 

\STATE {{\quad \quad \quad \quad $f_{\mathcal{C}_{1:T}} \longleftarrow $ ApplyPolicy($f_{\mathcal{C}_{m}}$, $f_{\mathcal{C}_{1:T}}$, $\rho$)} 
\textcolor{gray}{//Could return from 0 up to $T$ peers, here we assume all peers passed}}

\STATE {{\quad \quad \quad if $\textbf{IsPeerAnonymization}$}} 

\STATE {{\quad \quad \quad \quad ${f}_{a} \longleftarrow $ AnonymizePeers($f_{\mathcal{C}_{1:T}}$)} \textcolor{gray}{// Eq.\ref{eq4}}} 

\STATE {{\quad \quad \quad \quad ${\Phi}_{m} \longleftarrow $ LocalTraining($f_{\mathcal{C}_{m}},f_{a}$)}} 

\STATE {{\quad \quad \quad else} \textcolor{gray}{// No Peer Anonymization}}

\STATE {{\quad \quad \quad \quad ${\Phi}_{m} \longleftarrow $ LocalTraining($f_{\mathcal{C}_{m}},f_{\mathcal{C}_{1:T}}$)}}

\STATE {{\quad \quad \textbf{else}} \textcolor{gray}{// No Peer learning, standard federated learning}}

\STATE {{\quad \quad \quad ${\Phi}_{m} \longleftarrow $ LocalTraining($f_{\mathcal{C}_{m}}$)}} 

\STATE {{\quad \textbf{end for}}}

\STATE {{\quad $\Phi_{G} \longleftarrow \frac{1}{n} \sum_{j=1}^{n} \Phi_{j}$ } \textcolor{gray}{// Update global weights \ie FedAvg}} 

\STATE {{\quad ${\mathcal{F}_{m}} \longleftarrow $ extract features vector for each client}} 

\STATE {{\quad $\mathcal{W}_{M\times M} \longleftarrow $ update the similarity matrix} \textcolor{gray}{// Eq.\ref{eq2}}} 

\STATE {\textbf{end for}}
\end{algorithmic}

\end{algorithm}

\section{Experiments and Results}
We test our method on skin dermoscopic images through a set of experiments. Before that, we show proof of concept results of our method and compare it with current SOTA in \verb!SSFL! for CIFAR10 and FMNIST in image classification tasks in section~\ref{poc}. \verb!FedPerl! outperforms the baselines at different settings. Next, in section~\ref{exp:SLR}, we compare skin image classification results of our method with the baselines. The results show that peer learning enhances the performance of the models, yet applying PA enhances the communication cost in addition to the performance. After that, we show and discuss how \verb!FedPerl! builds the communities in section~\ref{exp:BC}. The results show that \verb!FedPerl! clusters the clients into main communities and individual clients thanks to our similarity matrix. Besides, \verb!FedPerl! boosts the overall performance of communities while it has a different effect on the individual clients. Thus, in section~\ref{exp:CLR}, we comment on the impact of the peer learning on the individual clients. \verb!FedPerl! shows superiority and less sensitivity to a noisy client. Then, we dig more deeply and present the classification results for each class in section~\ref{exp:CR}.  Our method enhances the classification for the individual classes, \emph{e.g.} up to 10 times for the DF class. Further, to confirm our findings and for more validation, we present the results using different evaluation metrics in section~\ref{secAdditional}. Our method is more calibrated and shows superiority over the \verb!SSFL! in the area under ROC and Precision-Recall curves, risk-coverage curve, and reliably diagrams. The qualitative results are presented in the same section. In section~\ref{secUClient}, we propose a more challenging scenario in which the clients do not have any labeled data. The classification results show that \verb!FedPerl! still achieves the best performance with or without PA. We end this part of experiments by showing the ability of \verb!FedPerl! to generalize to an unseen clients in section~\ref{secGUC}. {In section~\ref{secFedIRM}, we conduct a comparison with \verb!FedIRM!, a SOTA \verb!SSFL! method in skin lesion classification, under a fourth scenario where we have few labeled clients. Both models achieve comparable results when participation rate (PR)$=30\%$, while our method shows a lower performance when PR$=100\%$.} Note that the previous results were obtained when utilizing a static learning policy. Yet, in the last part of our experiments, we show the results of our dynamic peer learning policy in section~\ref{secLP}. In general, the new policy outperforms the results from the earlier one, while at the same time, it is successfully boosting the performance of the individual clients. 
\paragraph{\noindent\textbf{Datasets}}
\begin{figure*}[t]
\centering
\includegraphics[height=0.26\textheight]{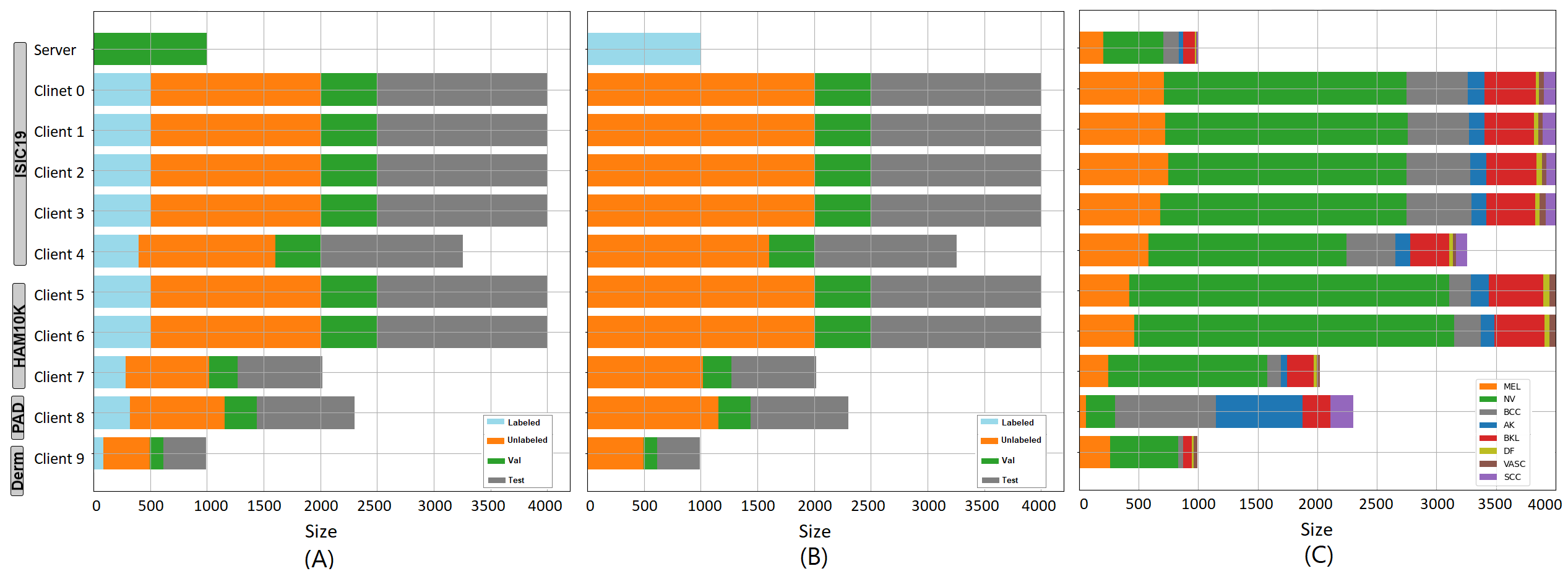}
\caption{Illustrative diagram shows the distribution of our clients. The datasets are divided randomly into ten clients besides the global model, without overlap between the clients. (A) The standard semi-supervised learning scenario. Each client data is divided into testing (gray), validation (green), labeled (blue), and unlabeled (orange) data. (B) The unlabeled/disjoint clients scenario. The labeled and unlabeled images are combined and used as unlabeled images. (C) The class distribution. The data split is designed to simulate a realistic scenario with severe class imbalance, varying data sizes, and diverse communities. The clients 5-9 missing one or more classes.}
\label{figdata}
\end{figure*}
Our database consists of 71,000 images collected from 5 publicly available datasets as the following. (1) ISIC19 \citep{codella2019skin} which consists of 25K images with 8 classes. The classes are melanoma (MEL), melanocytic nevus (NV), basal cell carcinoma (BCC), actinic keratosis (AK), benign keratosis (BKL), dermatofibroma (DF), the vascular lesion (VASC), and squamous cell carcinoma (SCC). (2) HAM10000 dataset \citep{tschandl2018ham10000} which consists of 10K images and includes 7 classes. (3) Derm7pt \citep{Kawahara2018-7pt} which consists of 1K images with 6 classes. (4) PAD-UFES \citep{pacheco2020pad} which consists of 2K images and includes 6 classes. The previous datasets are divided randomly into ten clients besides the global model, without overlap between datasets, \cf Fig.\ref{figdata}. (5) ISIC20 dataset \citep{rotemberg2021patient} which consists of more than 33K images with malignant ($\sim$ 500 images) and benign ($\sim$ 32.5K images) classes. The last dataset is used as testing data to study how \verb!FedPerl! generalizes to unseen data. Note that testing our method on the ISIC20 is a very challenging task due to the huge class imbalance and class distribution mismatch.   
\paragraph{\noindent\textbf{Baselines}}
We conduct our experiments on the following baselines; (i) Local models: which include lower, upper, and SSL (\verb!FixMatch! \citep{sohn2020fixmatch}) models, these models are trained on their local data without utilizing the federated learning. (ii) Federated learning models: which include lower, upper, and \verb!SSFLs! similar to \citep{yang2021federated}, and \verb!FedMatch!~\citep{jeong2020federated} models, where these models trained locally on their data and utilizing the federated learning globally. (iii) Ablation for our method, namely \verb!FedPerl! with(out) the PA. Note that for ease of implementation, we compare our method with one variant of \verb!FedMatch! that do not implement weights decomposition.
\paragraph{\noindent\textbf{Scenarios}}
\label{Scenarios}
Our experiments conducted under \tariq{four} scenarios. In the first scenario, the standard semi-supervised learning, \cf Fig.\ref{figdata}.(A), each client data is divided into testing (gray), validation (green), labeled (blue), and unlabeled (orange) data. The data split intended to resemble a realistic scenario with varying data size, severe class imbalance, and diverse communities, \textit{e.g.}, the clients 0-4 originated from ISIC19, the clients 5-7 originated from HAM10000, and client 8 and 9 originated from Derm7pt, and PAD-UFES, respectively. We train the lower bounds on the labeled data, while we train \verb!FixMatch!, \verb!SSFLs!, and \verb!FedPerl! on both labeled and unlabeled data. The upper bounds trained akin to SSLs, yet, all labels were exposed. In the second scenario, the unlabeled clients' scenario, we use the global data to train the global model. On the clients' side, however, the labeled and unlabeled images are combined and used as an unlabeled dataset, \ie the labels were excluded from the training, \cf Fig.\ref{figdata}.(B). While the second scenario is not yet investigated thoroughly in the medical images, we address it in this paper. In the third scenario, we test the ability of our model and the baselines to generalize to an unseen client (ISIC20) with new classes that have never been seen in the training. \tariq{The fourth scenario proposed by \citep{liu2021federated} in which there are few labeled clients. For this scenario, clients 1 \& 9 are selected as labeled clients while the remaining are not, such that they represent the largest community and individual clients, respectively.}    
\paragraph{\noindent\textbf{Implementation Details}}
We opt for EffecientNet \citep{tan2019efficientnet} pre-trained on ImageNet\citep{russakovsky2015imagenet} as a backbone architecture and trained using Adam optimizer~\citep{kingma2014adam} for 500 rounds. We follow FedVC~\citep{hsu2020federated} approach for clients federated learning. The idea of FedVC is to conceptually split large clients into multiple smaller ones, and repeat small clients multiple times such that all virtual clients are of similar sizes. Practically, this is achieved by fixing the number of training examples used for federated learning round to be fixed for every client, resulting in exactly optimization steps. The batch size $B$ and participation rate $(PR)$ were set to 16 \& 30\% (3 clients each round), respectively. The local training is performed for one epoch. The learning rate investigated in $[0.00001, 0.0001]$ and found best at $0.00005$. $\tau$ investigated in $[0.5, 0.95]$, and found best at $0.6$ $\&$ $0.9$ for the federated and local models respectively. $\beta$ investigated in $[0.1, 5]$, and found best at $0.5$. $\gamma$ investigated in $[0.01, 0.1]$, and found best at $0.01$. $T$ investigated in $\{2, 3, 4, 5\}$, and found best at $T=2$. The dynamic learning policy threshold $\rho$ tested at three values 0.75, 0.85, and 0.95, respectively. All images were resized to $224 \times 224$, and normalized to intensity values of $[0, 1]$. Random flipping and rotation were considered as weak augmentations, whereas RandAugment \citep{cubuk2020randaugment} was used as strong augmentation. We opt for PyTorch framework for the implementation hosted on standalone NVIDIA Titan Xp 12 GB machine. {As the followed procedures in semi-supervised learning, \verb!FedPerl! starts with warm-up rounds, e.g. 10 rounds in our case.} The testing results are reported for the models with best validation accuracy. The average training time takes around 7 hours for each run for \verb!FedPerl! models (w/o PA), about 5.85 hours for \verb!FedPerl! (with PA), about 5.5 hours for \verb!SSFL!, and about 6.25 hours for \verb!FedMatch! shedding the light on the cost effectiveness of our approach. \tariq{All the hyperparameters tuning was performed on a validation detest. Also, we made our code publicly available at \url{https://github.com/tbdair/FedPerlV1.0}.} 
\paragraph{\noindent\textbf{Evaluation Metrics}} 
We report the statistical summary of precision, recall, and F1-score. A Relative Improvement (RI) \emph{w.r.t} the baseline is also reported, where RI of $a$ over $b$ is $:(a-b)/b$. To highlight more in the model's performance at various threshold settings, we plot Area Under Receiver Operating Characteristic (AUROC) and Area Under Precision-Recall (AUPR) curves. Note that we follow the One vs ALL methodology for plotting. AUROC shows the model's ability to discriminate between positive examples and negative examples assuming balance data. Yet, AUPRC is a useful performance metric for imbalanced data, such as our case, where we care about finding positive examples.
Further, we investigate on the uncertainty evaluation and models confidence. Thus, we report Risk-Coverage (RC) curve \citep{geifman2017selective}, Reliability Diagram (RD) \citep{guo2017calibration}, and Expected and Maximum Calibration errors \citep{ding2020revisiting}, denoted as ECE and MCE respectively. RC curve plots the risk as a function of the coverage. The coverage denotes the percentage of the input processed by the model without rejection, while the risk denotes the level of
risk of the model's prediction \citep{geifman2017selective}. For a selective model, the mode abstains the prediction of input sample $x$ if the prediction confidence of that sample below a specific threshold e.g. 0.5. The higher coverage with lower risk, the better the model is. We refer the readers to section 2 in \citep{geifman2017selective} for the full definition of RC curve. Reliability Diagram, on the other hand, plots the accuracy as a function of confidence such that in the ideal case \ie a perfect calibrated model, the RD will plot the identity function.  For instance, suppose that we have 1000 samples, each with 0.85 confidence, we expect that 850 samples should be correctly classified. RD divides the predictions into different bins of confidence, \ie $B_{v}; v \in \{1, ..., V\}$, where $V$ is the total number of bins. Then, the average accuracy and the confidence for each bin $B_{v}$ are calculated as $acc(B_{v}) = \frac{1}{|B_{v}|} \sum_{i \in B_{v}} \textbf{1}(\Tilde{y_{i}} = y_{i})$, $conf(B_{v}) = \frac{1}{|B_{v}|} \sum_{i \in B_{v}} \Tilde{p_{i}}$, respectively, where $\Tilde{y_{i}}$, $y_{i}$, and $\Tilde{p_{i}}$ are the prediction, ground truth, and the confidence for sample $i$, respectively. The difference (gab) between the accuracy and the confidence can be positive when the confidence is higher than the accuracy, and negative when the accuracy is higher than the confidence. These gabs shown in the RD using different colors, \cf sec.\ref{secRD} and Fig.\ref{figRD}. 
For a perfect calibrated model, $acc(B_{v})=conf(B_{v})$ for all $v \in \{1, ..., V\}$. However, achieving a perfect calibrated model is impossible \citep{guo2017calibration}. 
Likewise, ECE and MCE are calculated, where ECE is defined as the difference in the weighted average of the bins' accuracy and confidence, while MCE represents the maximum difference, see Eq.\ref{eqECE} and Eq.\ref{eqMCE} respectively. 
\begin{align} 
\label{eqECE}
ECE = \sum_{v=1}^{V} \frac{|B_{v}|}{s} \Bigg| acc(B_{v})- conf(B_{v}) \Bigg|,
\end{align}
\begin{align} 
\label{eqMCE}
MCE = \max_{v \in \{1,..., V\}} \Big| acc(B_{v})- conf(B_{v})\Big|,
\end{align}
where $s$ is the number of samples in bin $B_{v}$. For a perfect calibrated model ECE and MCE both equal 0.
To calculate the reliability diagrams and calibration errors, we adopted an adaptive binning strategy~\citep{ding2020revisiting} that depends on fixable intervals in the calculations. This strategy is more accurate than using fixed intervals~\citep{ding2020revisiting}. Practically, we can realize the intervals used from the figure itself. For example, the width of the bars in figures \ref{figRD} and \ref{figRD20} represents the ranges used to calculate ECE and MCE.
\subsection{Proof-Of-Concept}
\label{poc}
First, we show proof of concept of our method on CIFAR-10 and FMNIST datasets and compare it with \verb!FedMatch!~\citep{jeong2020federated}; a  very recent work of \verb!SSFL!.  To have a fair comparison, we follow the codebases and the experimental setup they used. The results, reported in Table~\ref{tabelFedMatchDetails}, show that \verb!FedPerl! outperforms \verb!FedMatch!~\citep{jeong2020federated} in all experiments setup indicating the effectiveness of our method on finding the similarity (Sec~\ref{method:BC}), without introducing extra complexity, e.g., weight decomposition~\citep{jeong2020federated}.  Additionally, our \textit{peers anonymization} (PA) improves the accuracy and privacy at a low communication cost. Note that PA employs one anonymized peer while \verb!FedMatch! uses two clients in the training. {Interestingly, the FMNIST dataset results show that our method outperforms \verb!FedProx-SL!, which is inconsistent with the CIFAR10 dataset results. Although, these results are not comparable because \verb!FedProx-SL! results were taken from the original paper, whereas \verb!FedPerl! results were generated by our environment. Yet, this could be attributed to the fact that FMNIST images are much simpler than the ones in CIFAR10 yielding more robust similarities and hence producing more accurate pseudo labels.}
\begin{table}[!h] 
\caption{The classification accuracy for the Proof-Of-Concept experiment on CIFAR10 \& FMNIST datasets. *: Results as in FedMatch \citep{jeong2020federated}. SL: Fully supervised. The SSL methods use 10\% of the labeled data.}
\label{tabelFedMatchDetails}
\centering
\resizebox{1\textwidth}{!}{
\begin{tabular}{l|l||c|c|c}
\hline
PA& Method (SSFL)&CIFAR10 IID&CIFAR10 NonIID&FMNIST NonIID\\
\hline
\hline
* &FedAvg-SL & 58.60$\pm$0.42&55.15$\pm$0.21&-\\
*& FedProx-SL &59.30$\pm$0.31&57.75$\pm$0.15&82.06$\pm$0.26\\
*& FedAvg-UDA &46.35$\pm$0.29&44.35$\pm$0.39&-\\
*& FedProx-UDA  &47.45$\pm$0.21&46.31$\pm$0.63&73.71$\pm$0.17\\
*& FedAvg-FixMatch &47.01$\pm$0.43&46.20$\pm$0.52&-\\
*& FedProx-FixMatch &47.20$\pm$0.12&45.55$\pm$0.63&62.40$\pm$0.43\\
*& FedMatch &52.13$\pm$0.34&52.25$\pm$0.81&77.95$\pm$0.14\\
\hline
w/o PA & FedMatch (Our run) &53.12$\pm$0.65&53.10$\pm$0.99&76.48$\pm$0.18\\
w/ PA &FedMatch (Our run) &\textbf{53.32$\pm$0.59}&\textbf{53.80$\pm$0.39}&\textbf{76.72$\pm$0.44}\\
w/o PA&FedPerl      &\textbf{53.37$\pm$0.11}&\textbf{53.75$\pm$0.40}&\textbf{76.52$\pm$0.08}\\
w/ PA&FedPerl     &\textbf{53.98$\pm$0.06}&\textbf{53.50$\pm$0.71}&\textbf{82.75$\pm$0.44}\\
\hline 
\end{tabular}
}
\end{table}

\begin{figure*}[!t]
\centering
\includegraphics[width=\textwidth]{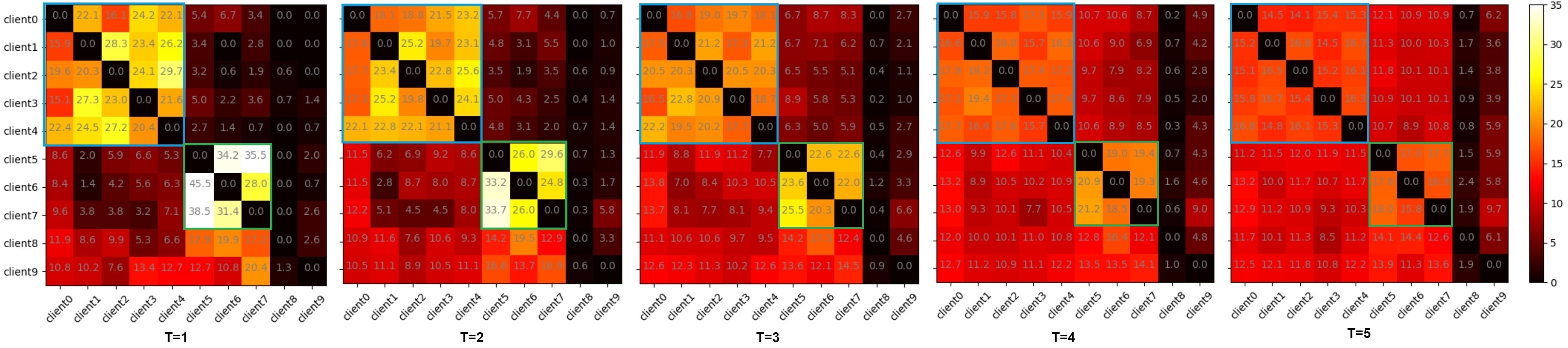}
\caption{FedPerl clusters clients into two main communities (blue \& green rectangles), while clients 8 \& 9 do not belong to any community. As we increase the committee size $T$, the frequency of selecting peers within the same community decreases. The numbers and colors correspond to the frequency, where the brighter colors or higher numbers values represent higher frequencies.}
\label{figcommunities}
\end{figure*}
\subsection{Skin Lesion Results}
\label{exp:SLR}
\begin{figure*}[!t]
\centering
\includegraphics[width=\textwidth]{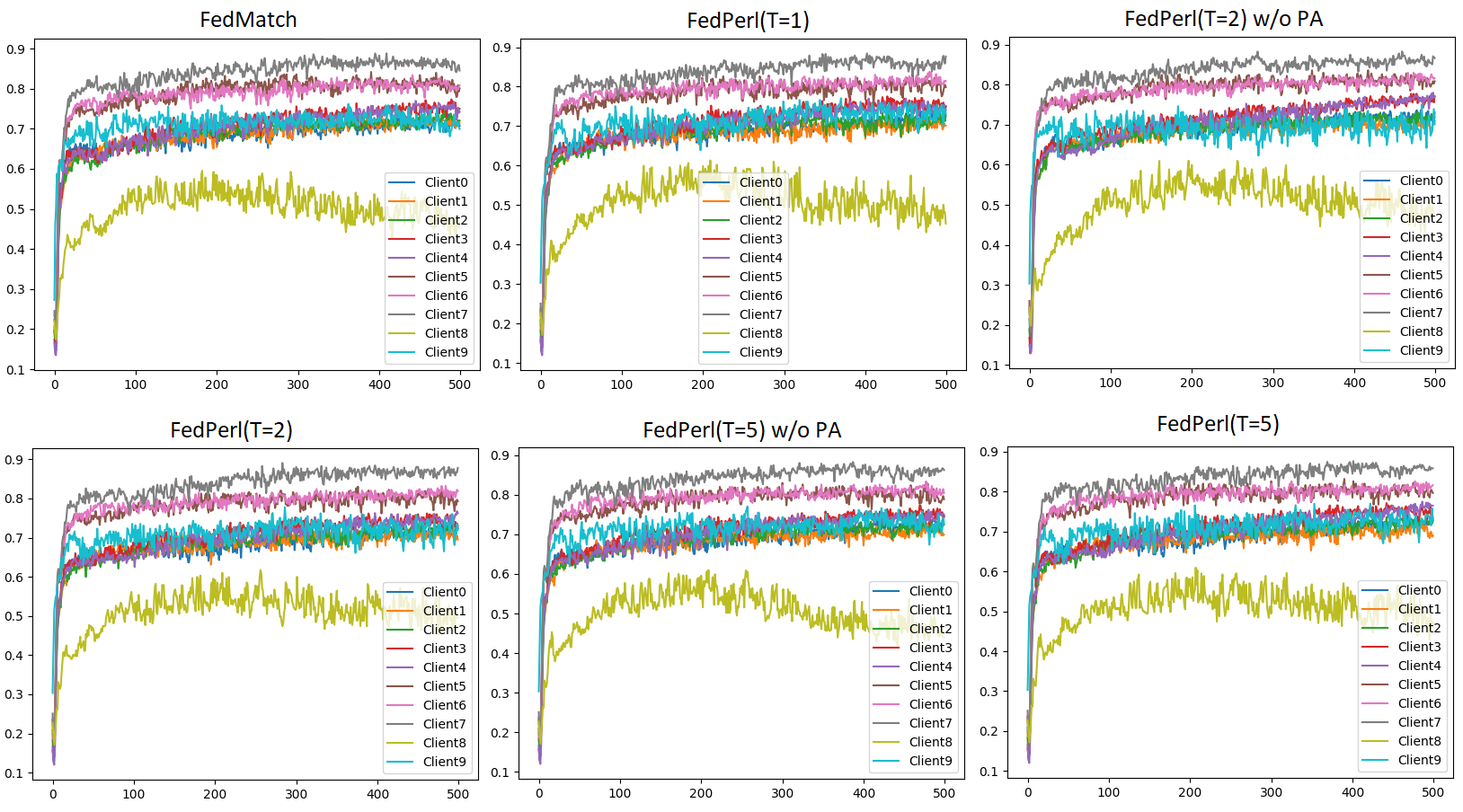}
\caption{The accuracy performance during the training. Due to the large number of curves that can be presented, we opt for FedMatch and FedPerl at different community sizes. The similar clients have achieved similar training performance.}
\label{figAccLog}
\end{figure*}

\begin{table*}[!t] 
\caption{The results under the standard semi-supervised learning scenario. Mean (Median) $\pm$ Std. of different evaluation metrics. $\dagger$:$\sim$\cite{yang2021federated}. $\ddagger$:$\sim$ FedMatch\citep{jeong2020federated}. RI: Relative Improvement. AC: Additional Cost. The AC is calculated \textit{w.r.t} the baseline (SSFL). For simplicity, we assume the initial cost for the SSFL is 0\%. +: with PA.}
\label{table1}
\centering
\resizebox{1\textwidth}{!}{
\begin{tabular}{l|l|c c c c c}
\hline
Setting & Model & F1-score & Precision & Recall &RI(\%)&AC(\%)\\
\hline
\hline
Lower   & Local   &0.647(0.632)$\pm$0.053 &0.644(0.622)$\pm$0.053&0.666(0.650)$\pm$0.053&-\\
              & FedAvg  &0.698(0.690)$\pm$0.084 &0.711(0.702)$\pm$0.072&0.709(0.700)$\pm$0.077&7.88\\
\hline
SSL     & FixMatch   &0.664(0.636)$\pm$0.060 &0.666(0.645)$\pm$0.063&0.692(0.671)$\pm$0.052&2.63&\\
SSFL   &FedAvg$\dagger$ &0.734(0.725)$\pm$0.065&0.744(0.730)$\pm$0.064&0.739(0.728)$\pm$0.061&13.44&0\\
&FedMatch$\ddagger$ &0.739(0.729)$\pm$0.076&0.751(0.745)$\pm$0.068&0.744(0.732)$\pm$0.071&14.22&200\\
w/o PA            & FedPerl(T=1) &\textbf{0.746(0.741)$\pm$0.071}&\textbf{0.753(0.744)$\pm$0.069}&\textbf{0.748(0.744)$\pm$0.069}&\textbf{15.30}&100\\
w/o PA             & FedPerl(T=2) &\textbf{0.747(0.736)$\pm$0.071} &\textbf{0.756(0.741)$\pm$0.067}&\textbf{0.750(0.739)$\pm$0.069}&\textbf{15.46}&200\\
w/o PA               & FedPerl(T=3)  &\textbf{0.746(0.741)$\pm$0.072}&\textbf{0.757(0.743)$\pm$0.066}&\textbf{0.747(0.743)$\pm$0.070}&\textbf{15.30}&300\\
w/o PA               & FedPerl(T=4)  &\textbf{0.741(0.731)$\pm$0.077}&\textbf{0.751(0.735)$\pm$0.069}&\textbf{0.745(0.736)$\pm$0.072}&\textbf{14.53}&400\\
w/o PA              & FedPerl(T=5) &\textbf{0.744(0.734)$\pm$0.073} &\textbf{0.753(0.744)$\pm$0.071}&\textbf{0.747(0.739)$\pm$0.069}&\textbf{15.00}&500\\
\hline
&FedMatch$+\ddagger$&\textbf{0.745(0.737)$\pm$0.071}&\textbf{0.750(0.737)$\pm$0.067}&\textbf{0.750(0.746)$\pm$0.069}&\textbf{15.15}&100\\           
    & FedPerl(T=2) &\textbf{0.746(0.737)$\pm$0.075} &\textbf{0.754(0.741)$\pm$0.071}&\textbf{0.749(0.742)$\pm$0.073}&\textbf{15.30}&100\\
    & FedPerl(T=3)  &\textbf{0.746(0.738)$\pm$0.066}& \textbf{0.756(0.743)$\pm$0.060}&\textbf{0.748(0.740)$\pm$0.065}&\textbf{15.30}&100\\
& FedPerl(T=4)  &\textbf{0.746(0.736)$\pm$0.077} &\textbf{0.755(0.745)$\pm$0.072}&\textbf{0.750(0.740)$\pm$0.074}&\textbf{15.30}&100\\
& FedPerl(T=5) &\textbf{0.749(0.739)$\pm$0.068}&\textbf{0.758(0.744)$\pm$0.065}&\textbf{0.750(0.742)$\pm$0.066}&\textbf{15.77}&100\\
\hline
Upper   & Local   &0.726(0.701)$\pm$0.044 &0.729(0.705)$\pm$0.045&0.732(0.710)$\pm$0.042&12.21\\
              & FedAvg  &0.773(0.757)$\pm$0.068 &0.779(0.765)$\pm$0.065&0.773(0.759)$\pm$0.069&19.47\\
\hline 
\end{tabular}
}
\end{table*}
\paragraph{\noindent\textbf{Federated Learning Results}}
In this section, we present the federated learning classification results before applying our method \ie without peer learning nor PA. The results in Table~\ref{table1} proves the current findings that \verb!FedAvg! outperforms the local models significantly.  For example, see Local/\verb!FixMatch! vs. \verb!FedAvg!, the obtained F1-score are 0.647 and 0.698, 0.664 and 0.734, and 0.726 and 0.773, respectively, with relative improvement (RI) up to $19.74\%$. Interestingly, both lower \verb!FedAvg! and \verb!FedAvg! $\ddagger$ (\verb!SSFL!) models exceed the local SSL and upper bound models, respectively.  That implies aggregating knowledge across different clients is more beneficial than exploiting local unlabeled or labeled data individually. Next, we discuss \verb!FedPerl! results at different values of $T$.
\paragraph{\noindent\textbf{FedPerl results without PA}}
\label{exp:PLR}
The results of  \verb!FedPerl! without applying peer anonymization is shown in Table \ref{table1} (denoted as w/o PA). The first concluding remarks reveal that peer learning enhances the local models. For illustration, our method outperforms the lower model with RI between $14.53\%$ and $15.46\%$. Further, \verb!FedPerl! exceeds (\verb!SSFL!) \verb!FedAvg!$\dagger$~\citep{yang2021federated} and \verb!FedMatch!~\citep{jeong2020federated} by $1.8\%$ and $1.08\%$, respectively. Moreover, our approach better than the local upper bound by $2.9\%$. Note that \verb!SSFL! is considered a special case of \verb!FedPerl! when $T=0$. 
In addition, \verb!FedPerl! results at a different number of peers $T$ (committee size) are comparable, while the communication cost, comparing to the standard \verb!SSFL!, increases proportionally with the increasing value of $T$ (see AC in Table \ref{table1}). Note that, the additional cost is calculated with respect to the baseline (\verb!SSFL!). For simplicity, we assume the initial cost for the \verb!SSFL! is 0\%. Finally, the results imply that employing one similar peer ($T=1$) is adequate to obtain remarkable enhancement with minimal communication cost, yet, at the loss of privacy. To address this, we propose \textbf{peers anonymization} technique.  
\paragraph{\noindent\textbf{FedPerl results}} 
\label{exp:PLRPA}
After applying the peer anonymization, all models show a similar or slightly better performance when compared to the previous results (\ie  w/o PA), \cf Table \ref{table1}. Yet, the new models enhance the baseline's performance, while still being better at hiding clients' identities and reducing the communication cost $O(1)$ regardless of the committee size $T$. 
Interestingly, applying peer anonymization not only enhances \verb!FedPerl!, but also the \verb!FedMatch! method. Specifically, the F1-score increases from 0.739 to 0.745, see \verb!FedMatch! vs. \verb!FedMatch+! in Table~\ref{table1}. Note that the additional advantages of \verb!FedMatch!+ over \verb!FedMatch! are the anonymized peer and the communication cost.
The improvement of performance is attributed to the carefully designed strategy of creating the anonymized peer, such that the learned knowledge from many models ensembled into a single model. The results confirm the superiority of \verb!FedPerl!, and show that our peer anonymization is orthogonal and can be easily integrated into other methods without additional complexity. 
\tariq{
In Fig.~\ref{figAccLog}, we show the accuracy performance during the training. While we notice that similar clients have achieved similar training behavior, no further improvement in the last stages of the training was observed for all clients. For example, the accuracy for the clients 0-4 between 75-65, while it is between 80-90 for the clients 5-7. Client 9 achieved accuracy that is similar to clients 0-4. However,  the best accuracy for client 8 was achieved in the middle of the training. This suggests handling Out-of-Distribution clients in federated learning has to be further investigated.}
\subsection{Building Communities Results}
\label{exp:BC}
In this experiment, we investigate the importance of the similarity matrix used to rank similar clients and cluster them into communities. In Fig.\ref{figcommunities} we present the percentage of selecting peers during the training at different $T$ values. To gain more insights, let us consider when the community size (T=2).  For instance, the percentage of $33.2\%$ between clients 6 and 5 reflects how often client 5 has chosen as a similar peer for client 6. The blue \& green rectangles show that the clients clustered into two main communities.  Interestingly, the clustering matches the clients' distribution we designed in our experiment, \cf Fig.\ref{figdata}.  For further analysis on community 1 (blue rectangle), we find the frequency of selecting peers from the same community for each client by calculating the horizontal summations (columns 0-4). The frequencies are $81.6\%, 85.6\%, 89.5\%, 86.4\%$, and $88.1\%$ for the clients 0-4, respectively. That suggest, for example, client 0 learns from its community with a percentage of $81.6\%$ of the training time. On average $86.24\%$ of the time, first community members learn from each other, while it is $57.77\%$ for community 2 (green rectangle).  
The same clustering also is shown for \verb!FedPerl! at different committee sizes; $T=\{1, 3, 4, 5\}$. Note that the frequency values are gradually decreasing when a larger committee size is used for communities 1 \& 2. The decrease in frequencies is expected because the likelihood of selecting peers from the outside of the community increases as we use a bigger committee size. Hence, the frequencies are distributed among the clients. In contrast, the frequencies for selecting peers for the individual clients (8 \& 9) are comparable to each other at different $T$ values.
\begin{table}[!h] 
\caption{The mean F1-score is reported to show the influence of peer learning on the community and individual clients. M: number of clients. *: SSFL.}
\label{tableCommu}
\centering
\resizebox{0.70\textwidth}{!}{
\begin{tabular}{l|c c c c}
\hline
Model&$C_{ISIC}$&$C_{HAM}$& 8 & 9\\
& (M=5)& (M=3) & (M=1)& (M=1)\\
\hline
\hline
FedPerl(T=0)* &0.718&0.816&0.602&0.703\\
FedPerl(T=1) &0.738&0.829&0.584&0.699\\
FedPerl(T=2) &0.736&0.833&0.567&0.717\\
FedPerl(T=3) &0.735&0.828&0.594&0.725\\
FedPerl(T=4) &0.735&0.826&0.562&0.727\\
FedPerl(T=5) &0.737&0.824&0.588&0.731\\
\hline
\end{tabular}
}
\end{table}
For further analysis on the community results, we average the classification results in each community and report them in Table~\ref{tableCommu}. The first note from the results indicates that peer learning boosts the overall performance of the communities, compare the values of $T=0$ vs. $T=\{1, 2, 3,4, 5\}$ for $C_{ISIC}$ and $C_{HAM}$ respectively. Note that peer learning is not applied when $T=0$.
Further, we notice a stable performance for the $C_{ISIC}$ community after applying the peer learning regardless of $T$ values, yet with slight changes. However, an increasing then a decreasing performance is observed for the $C_{HAM}$ at increasing values of $T$. 
This performance inconsistency is attributed to the community size. For instance, $C_{ISIC}$ community includes 5 clients, while $C_{HAM}$ community contains 3 clients. At first, let us consider $C_{ISIC}$. The probability of selecting peers, based on their similarities,  from the outside community for different values of $T\geq1$ is very low, and most likely the peers coming from the same community \ie internal peers. For the case when $T=5$, selecting an external peer is guaranteed. Yet, its effect is negligible comparing to the other clients, who most likely are internal peers. 
Now let us consider $C_{HAM}$. We notice that the performance increases gradually and reaches the best at $T=2$. Based on our similarity matrix, the peers until this value most likely are internal peers,  yielding to enhancement in the performance. Yet, after that (\ie $T > 2$), involving external peers is confirmed. Consequently, the local model is distracted by increasing the number of external peers when using larger $T$ values. Hence, the decrease in the performance.  
On the other hand, the individual clients' results are interesting (\ie clients 8 \& 9). While an enhancement is noticed for client 9, a reduction is observed for client 8. We note that the accuracy of client 9 is increased as the committee size increases thanks to peer learning. In general, the large the committee size, the better the performance. Yet, peer learning harms client 8. One explanation is attributed to the class distribution mismatch between client 8 and the other clients, \cf Fig.\ref{figdata}.(C). Further analysis is discussed in the next section concerning the individual clients' performance.       

\paragraph{\noindent\textbf{Random peers}}
\label{rcc}
To investigate the importance of peer learning and our similarity matrix, we perform an additional experiment where the peers for the clients are selected randomly. The obtained F1-score is $0.736$, with RI equals $13.75\%$ and $0.27\%$ \emph{w.r.t.} the lower bound and \verb!SSFL!, respectively. These results imply two conclusions. (i) Even with random clients, peer learning is still beneficial to training, compare this experiment results with the \verb!SSFL!. (ii) Utilizing our similarity matrix brings extra knowledge by picking more accurate peers, compare this experiment results with the \verb!FedMatch! models.            
\subsection{The Influence of Peer Learning on Clients}
\label{exp:CLR}
\begin{table}[!t] 
\caption{The classification results for each client (mean F1-score). $C_{ISIC}$, $C_{HAM}$: the results at the community level. $Avg_{/C8}$: the results after excluding client 8.  $\dagger$:$\sim$\cite{yang2021federated}. $\ddagger$:$\sim$FedMatch\citep{jeong2020federated}. Diff.\% = $Avg_{/C8}$ - $Avg$. $+$: with PA.}
\label{table2}
\centering
\resizebox{\textwidth}{!}{
\begin{tabular}{l|l|c c c c c |c|c c c|c|c c |c|c|c}
\hline
Setting&Model/Client& 0& 1& 2& 3& 4&$C_{ISIC}$ & 5& 6& 7&$C_{HAM}$& 8& 9 & $Avg_{/C8}$ & Avg &Diff.\%\\
\hline
\hline
Lower &Local  &0.581&0.618&0.603&0.622&0.596&0.604&0.742&0.738&0.670&0.717&0.656&0.641&0.646$\pm$0.056&0.647$\pm$0.053&-\\
      &FedAvg &0.678&0.687&0.667&0.703&0.692&0.685&0.794&0.796&0.787&0.792&0.492&0.684&0.731$\pm$0.047&0.698$\pm$0.084&3.3\\
\hline
SSL &FixMatch&0.634&0.635&0.608&0.637&0.626&0.628&0.752&0.783&0.716&0.751&\textbf{0.650}&0.602&0.666$\pm$0.063&0.664$\pm$0.060&-\\
SSFL    & FedAvg $\dagger$ &0.727&0.718&0.686&0.723&0.735&0.718&0.812&0.831&0.806&0.816&0.602&0.703&0.764$\pm$0.045&0.734$\pm$0.065&3.0\\
   & FedMatch$\ddagger$&0.724&0.724&0.722&0.734&0.751&0.731&0.801&0.850&0.813&0.822&0.553&0.717&0.760$\pm$0.046&0.739$\pm$0.076&2.1\\
    & FedMatch+$\ddagger$&\textbf{0.733}&\textbf{0.740}&\textbf{0.729}&\textbf{0.734}&0.744&\textbf{0.736}&\textbf{0.813}&0.843&\textbf{0.826}&\textbf{0.827}&0.581&0.703&\textbf{0.768}$\pm$\textbf{0.053}&\textbf{0.745}$\pm$\textbf{0.071}&2.3\\
w/o PA  & FedPerl(T=2) &\textbf{0.735}&\textbf{0.731}&\textbf{0.725}&\textbf{0.737}&0.739&\textbf{0.733}&0.805&\textbf{0.850}&\textbf{0.839}&\textbf{0.831}&0.582&\textbf{0.729}&\textbf{0.769}$\pm$\textbf{0.047}&\textbf{0.747}$\pm$\textbf{0.071}&2.2\\
     & FedPerl(T=2) &\textbf{0.737}&\textbf{0.737}&\textbf{0.724}&0.730&\textbf{0.751}&\textbf{0.736}&\textbf{0.818}&0.846&\textbf{0.834}&\textbf{0.833}&0.567&\textbf{0.717}&\textbf{0.765}$\pm$\textbf{0.046}&\textbf{0.746}$\pm$\textbf{0.075}&1.9\\
\hline
Upper & Local  &0.698&0.698&0.677&0.700&0.696&0.694&0.806&0.804&0.752&0.787&0.702&0.722&0.728$\pm$0.046&0.726$\pm$0.044&-\\
& FedAvg &0.736&0.747&0.735&0.753&0.761&0.746&0.859&0.855&0.861&0.858&0.630&0.789&0.797$\pm$0.054&0.773$\pm$0.068&2.4\\
\hline
\end{tabular}
}
\end{table}

This experiment aims to gain more insights on the individual results and realize the influence of peer learning on clients and compare it with the baselines. The results are shown in Table \ref{table2}. We observe that \verb!FedPerl! exceeds the baselines, including the local upper bounds with salient margins, \emph{e.g.}, for client 7 it is about $16.4\%$ (Lower Local vs. \verb!FedPerl!). 
In the same direction, \verb!FedPerl! steadily surpasses \verb!FedMatch! at the community's level and in all individual clients' results except for client 4. Yet, thanks to our PA, \verb!FedMatch+! shows better results than \verb!FedMatch! at all communities and clients except for clients 4, 6, and 9.
Surprisingly, \verb!FedPerl! excels the upper \verb!FedAvg! for client 0. The performance improvement is observed for all clients except client 8. One explanation is that \verb!FedPerl! does not find suitable peers for client 8 to learn from due to the class distribution mismatch (\cf Fig.~\ref{figdata}.(C)).
For further investigation on the impact of client 8, we explore excluding it from the training. Then we compare the new and the previous results, both reported as $Avg_{/C8}$ and $Avg$ respectively in Table \ref{table2}. The comparison unveils that all federated learning models (\ie \verb!FedAvg!, \verb!FedMatch!, and \verb!FedPerl!) obtain better performance after excluding client 8. Still, the best performance is observed for \verb!FedPerl! over the local upper and the (\verb!SSFL!) \verb!FedAvg! models. Note that the performance reduction after including client 8 in the training (see $Avg$ in Table \ref{table2}) implies the negative impact of this client. To realize that, we calculate the difference in performance before and after including client 8, \ie $Avg_{/C8} - Avg$, and report the results in column $Diff.$ in Table \ref{table2}. The resulted values show the negative impact of client 8 on the results. Where the higher the difference is, the higher the negative impact is. For example, it negatively impacted, the smallest on \verb!FedPerl! ($1.9\%$), moderate on \verb!FedPerl! w/o PA ($2.2\%$) and on both \verb!FedMatch! methods ($2.1\%$ and $2.3\%$), and the largest on \verb!FedAvg! ($3\%$).
Such negative behavior could represent a threat in the federated learning, where a noisy and out-of-distribution client might hurt other clients and mislead the global model. Yet, the most interesting observation from this experiment that \verb!FedPerl! is less prone to the negative and noisy impact than \verb!SSFLs!, thanks to the training schema we proposed.
We do not claim that \verb!FedPerl! is robust against class distribution mismatch, but rather it is less sensitive to a noisy client. Nevertheless, the inconsistency in behavior between clients 8 \& 9 could be further investigated. 

On the other side, we notice that the enhancement after applying peer learning also observed at the community level; $C_{ISIC}$ and $C_{HAM}$ with $13.2\%$ and $11.6\%$, respectively, confirming the finding in the previous section.

Note that our final objective consists of two terms that try to achieve the balance between the local and global benefits. Experimentally, we have shown that client 8 harms the clients. This impact was the minimum on \verb!FedPerl! who is utilizing peer learning. Thus, we argue that involving peers, who influence the local models through participating in the pseudo labeling,  has two advantages; (i) it restricts client 8 to send more reliable updates, and (ii) it reduces the negative influence of that client. Also, the $T$ peers learn and coach the local client and guide it to be more accurate, where a noisy client could be fixed by averaging with more reliable clients.
\subsection{Class Level Results}
\label{exp:CR}
\begin{table}[!t] 
\caption{The classification results for the eight classes (mean F1-score). +: with PA. $\dagger$:$\sim$\cite{yang2021federated}. $\ddagger$:$\sim$FedMatch\citep{jeong2020federated}}
\label{table4}
\centering
\resizebox{1\textwidth}{!}{
\begin{tabular}{l|l|c c c c c c c c}
\hline
Setting& Model& MEL & NV & BCC & AK & BKL & DF & VASC & SCC\\
\hline
\hline
Lower &Local   &0.430&0.811&0.502&0.293&0.357&0.099&0.318&0.124\\
&FedAvg &0.501&0.834&0.646&0.377&0.507&0.173&0.642&0.111\\
\hline
SSL&FixMatch&0.451&0.831&0.540&0.304&0.374&0.052&0.292&0.135\\
SSFL &FedAvg $\dagger$ &0.565&0.852&0.680&\textbf{0.396}&0.570&0.416&0.707&0.253\\
&FedMatch$\dagger$&0.573&0.852&0.700&0.366&0.565&0.462&0.701&0.275\\
&FedMatch+&\textbf{0.579}&\textbf{0.853}&\textbf{0.701}&0.376&\textbf{0.574}&\textbf{0.506}&\textbf{0.708}&\textbf{0.302}\\
w/o PA&FedPerl(T=2) &\textbf{0.576}&\textbf{0.854}&\textbf{0.706}&0.393&\textbf{0.589}&\textbf{0.552}&0.702&\textbf{0.305}\\
&FedPerl(T=2) &\textbf{0.602}&\textbf{0.854}&0.687&0.390&\textbf{0.592}&\textbf{0.493}&\textbf{0.712}&\textbf{0.315}\\
\hline
Upper& Local   &0.551&0.853&0.651&0.428&0.520&0.308&0.654&0.308\\
&FedAvg  &0.617&0.867&0.750&0.510&0.637&0.672&0.804&0.282\\
\hline 
\end{tabular}}
\end{table}

\begin{table}[!t] 
\caption{The area under ROC curve for the eight classes. +: with PA. $\dagger$:$\sim$\cite{yang2021federated}. $\ddagger$:$\sim$FedMatch\citep{jeong2020federated}}
\label{tableROC}
\centering
\resizebox{1\textwidth}{!}{
\begin{tabular}{l|l|c c c c c c c c}
\hline
Setting& Model& MEL & NV & BCC & AK & BKL & DF & VASC & SCC\\
\hline
\hline
Lower &Local   &0.662 &0.777 &0.760 &0.677 &0.644 &0.529 &0.676 &0.540\\
&FedAvg &0.709 &0.834 &0.827 &0.634 &0.739 &0.575 &0.909 &0.528\\
\hline
SSL&FixMatch &0.670 &0.804 &0.785 &0.685 &0.658 &0.515 &0.644 &0.540\\
SSFL &FedAvg$\dagger$    &0.737 &0.846 &0.827 &\textbf{0.692} &0.777 &0.675 &0.889 &0.583\\
&FedMatch$\ddagger$ &0.749 &0.851 &0.843 &0.650 &0.772 &0.690 &0.897&0.586\\
&FedMatch+   &\textbf{0.751} &\textbf{0.854} &\textbf{0.847} &0.655 &\textbf{0.778}&\textbf{0.717} &\textbf{0.904}&\textbf{0.596}\\
w/o PA&FedPerl(T=2)   &\textbf{0.750}&\textbf{0.859}&\textbf{0.853}&0.659 &\textbf{0.785} &\textbf{0.732} &\textbf{0.900} &\textbf{0.608}\\
&FedPerl(T=2)         &\textbf{0.758}&\textbf{0.860}&0.838&0.660 &\textbf{0.791} &\textbf{0.717} &\textbf{0.907} &\textbf{0.608}\\
\hline
Upper& Local   &0.728 &0.838 &0.831 &0.733 &0.733 &0.648 &0.824 &0.606\\
&FedAvg  &0.773 &0.869 &0.848 &0.750 &0.805 &0.876 &0.958 &0.594\\
\hline 
\end{tabular}}
\end{table}

Because our setting is heterogeneous and suffers from severe class imbalance (\cf Fig.\ref{figdata}.(C)), it is of importance to validate our method in that setting. Thus, we report the class level performance in Table \ref{table4}. \verb!FedPerl! obtains skin lesion classification accuracy better than local models (\verb!FedPerl! vs. Local/\verb!FixMatch!). For example, the improvement reaches ten times in the DF class. Moreover, \verb!FedPerl! enhances the accuracy for BCC, BKL, DF, VASC, and SCC lesions by $16.6\%$, $21.8\%$, $50.4\%$, $42.0\%$, and $18.0\%$, respectively, in the SSL setting. The comparison with \verb!FedMatch! reveals the same behavior seen in the previous results. First, our method, in general, outperforms \verb!FedMatch! in all lesions. Second, applying PA to \verb!FedMatch! (denoted as \verb!FedMatch+!) boosts its accuracy. On the other hand, we observe an insignificant decrease in the accuracy of the AK lesion. The key factor of \verb!FedPerl! advantage is attributed to the knowledge exchanged through peer learning.  
\subsection{Additional Evaluation Metrics}
\label{secAdditional}
\begin{table}[!t] 
\caption{The area under Precision-Recall curve for the eight classes. +: with PA. $\dagger$:$\sim$\cite{yang2021federated}. $\ddagger$:$\sim$FedMatch\citep{jeong2020federated}}
\label{tablePRC}
\centering
\resizebox{1\textwidth}{!}{
\begin{tabular}{l|l|c c c c c c c c}
\hline
Setting& Model& MEL & NV & BCC & AK & BKL & DF & VASC & SCC\\
\hline
\hline
Lower &Local   &0.505 &0.864 &0.622 &0.457 &0.409 &0.287 &0.524 &0.164\\
&FedAvg &0.582 &0.894 &0.702 &0.443 &0.556 &0.394 &0.712 &0.313\\
\hline
SSL&FixMatch &0.527 &0.879 &0.646 &0.476 &0.453 &0.358 &0.534 &0.216\\
SSFL &FedAvg$\dagger$    &0.620 &0.903 &0.730 &\textbf{0.494}&0.617 &0.574 &0.762 &0.348\\
&FedMatch$\ddagger$ &0.640 &0.906 &0.745 &0.460 &0.615 &0.559 &0.753 &0.344\\
&FedMatch+&\textbf{0.645} &\textbf{0.908} &\textbf{0.752}&0.479&\textbf{0.632}&\textbf{0.598} &0.761 &\textbf{0.349}\\
w/o PA&FedPerl(T=2)   &\textbf{0.642}&\textbf{0.910}&\textbf{0.751}&0.476 &\textbf{0.630}&\textbf{0.618} &0.754 &\textbf{0.368}\\
&FedPerl(T=2)         &\textbf{0.651}&\textbf{0.911}&0.744&0.475 &\textbf{0.629}&\textbf{0.627}&\textbf{0.769}&\textbf{0.356}\\
\hline
Upper& Local  &0.596 &0.899 &0.710 &0.561 &0.555 &0.456 &0.690 &0.361\\
&FedAvg  &0.668 &0.916 &0.762 &0.574 &0.670 &0.719 &0.847 &0.373\\
\hline 
\end{tabular}}
\end{table}

\begin{figure}[!t]
\centering
\includegraphics[width=\textwidth, height=0.95\textheight]{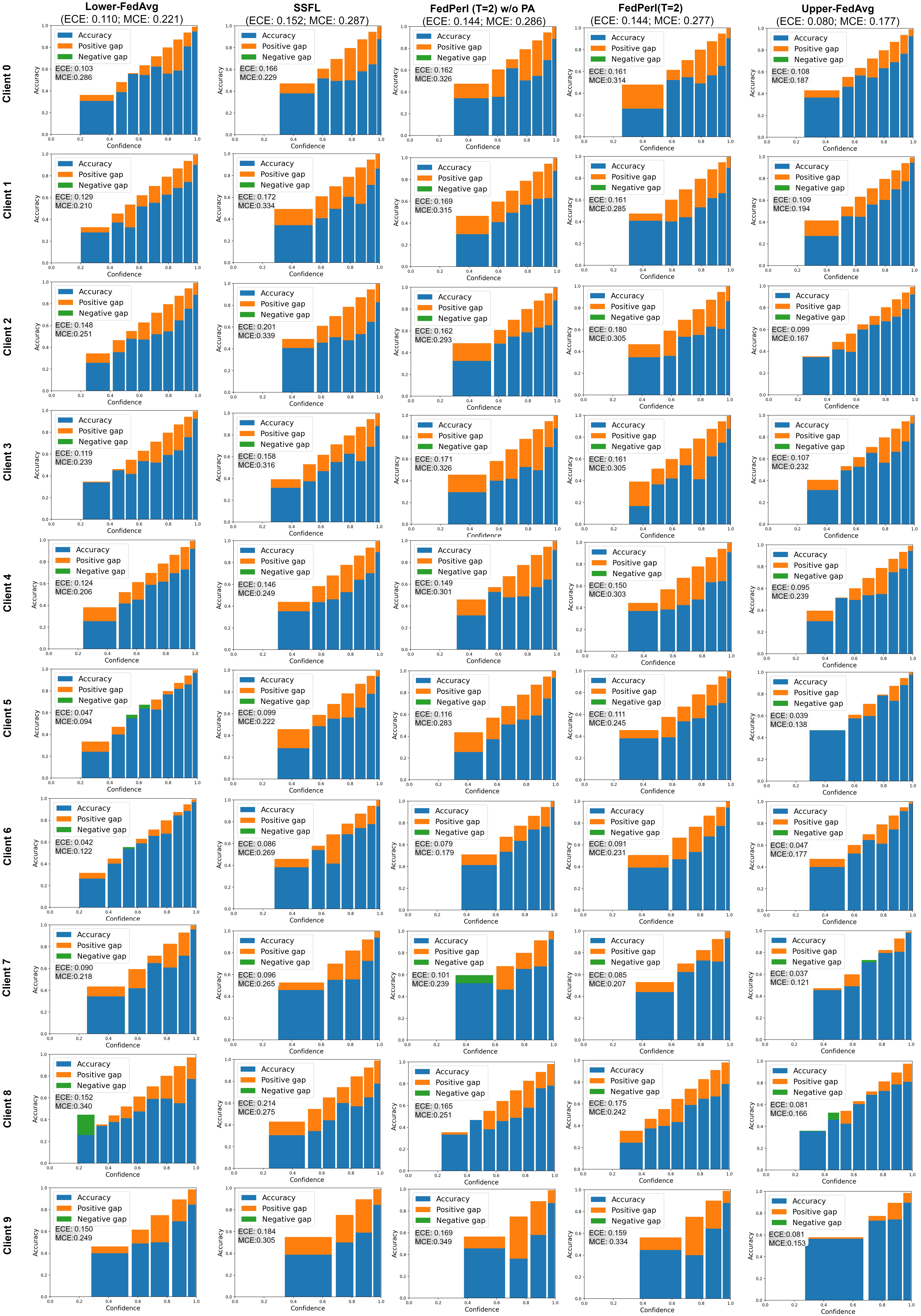}
\caption{Reliability diagrams and calibration errors. FedPerl is more calibrated than SSFL and local upper models indicating better and more confident predictions. The local models are shown in the supplementary materials.}
\label{figRD}
\end{figure}

\paragraph{\noindent\textbf{Area under ROC \& Precision-Recall curves}}
For more validation, we report the area under ROC curve (AUROC) and the area under Precision-Recall curve (AUPRC) in Table \ref{tableROC} and Table \ref{tablePRC} respectively. It is clearly shown that \verb!FedPerl! exceeds \verb!SSFLs! in all classes results except for the AK class. For instance, in AUROC results, the enhancement of \verb!FedPerl! over \verb!SSFL! around 2.1\%, 1.4\%, 1.1\%, 1.4\%, 4.2\%, 1.8\%, and 2.5\% for the MEL, NV, BCC, BKL, DF, VASC, and SCC classes respectively. Moreover, the boosting of \verb!FedPerl! over the lower \verb!FedAvg! reaches 14\% for the DF class. Interestingly, \verb!FedPerl! outperforms both upper bounds for the SCC class. On the other side, the comparison of the AUPRC results reveals the same observations.
Finally, the superiority of our method still found over \verb!FedMatch!. 
\paragraph{\noindent\textbf{Risk Coverage curve}}
\begin{figure*}[!t]
\centering
\includegraphics[width=\textwidth]{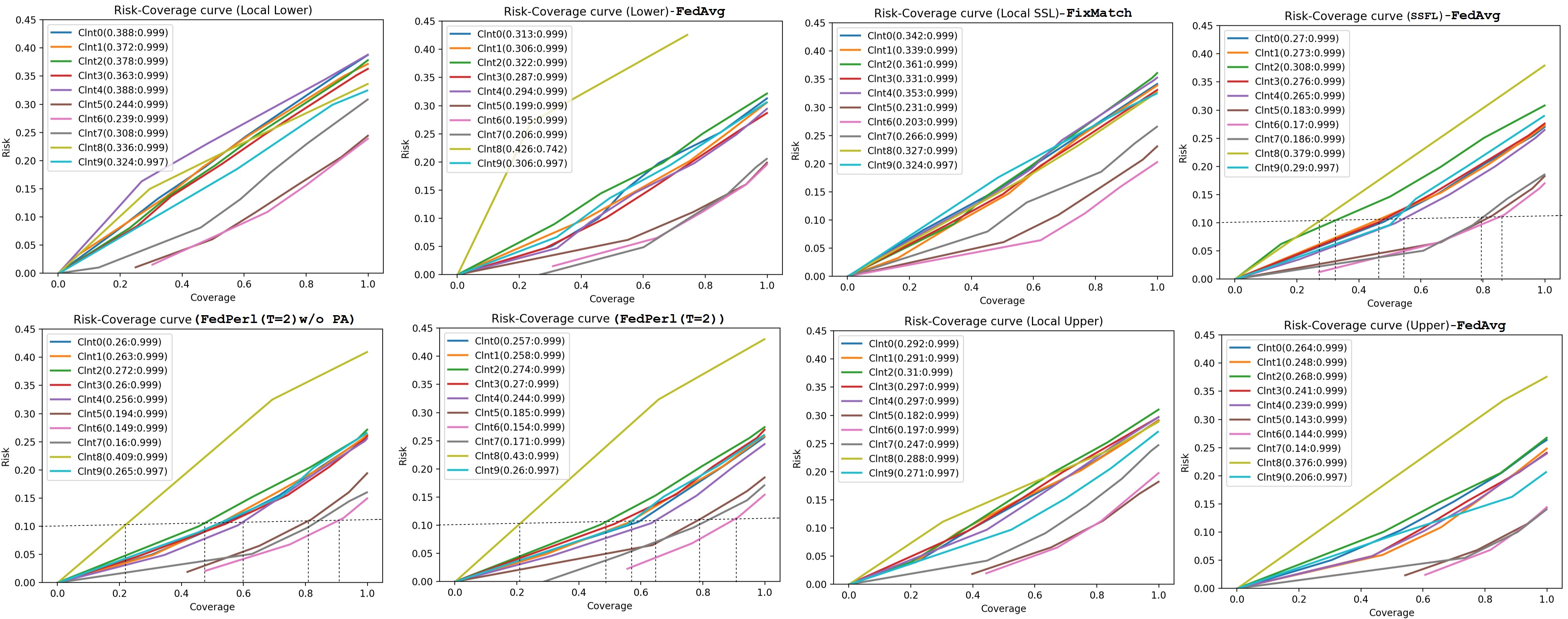}
\caption{The area under Risk-Coverage curve for FedPerl and the baselines. The numbers that appear next to a client name represent the risk at the full coverage respectively \ie (risk: coverage). In general, FedPerl obtains the lowest risk with the best coverage among all models.}
\label{figAURC}
\end{figure*}
We show the Risk-Coverage curves for \verb!FedPerl! and our baselines in Fig.\ref{figAURC}. Each plot in the figure depicts a model. Inside each plot, we draw the curves for all clients. The numbers that appear next to a client name represent the risk value at the full coverage of the input data, \ie (risk: coverage). It is shown from the figures that \verb!FedPerl! achieves the lowest risk with the best coverage amongst all models, and this for all clients except for clients 5 \& 8. Note that the coverage of client 8 in all federated models is worse than the local models, which is attributed to class mismatch. Please refer to sections \ref{exp:BC} and \ref{exp:CLR} for more details. Nevertheless, if we consider the clients 0, 4, and 9 as examples, we observe that \verb!FedPerl! obtains the maximum coverage at risks of 25.7\%, 24.4\%, and 26.0\% respectively. These values are better than all local models including the upper local model,  and better than \verb!FedAvg! SSL (\verb!SSFL!) model. Though, an insignificant drop in the coverage is noticed for client 5 comparing to \verb!SSFL!. A detailed comparison between \verb!FedPerl! and \verb!SSFL! at 10\% risk shows the superiority of \verb!FedPerl! over \verb!SSFL! in all clients, except client 8. For instance, the coverage jumps from $(33\%$ to $49\%)$ for client 2, from the range of $(46.5\%-55\%)$ to the range of $(53\%-65\%)$ for clients 0, 1, 3, 4, and 9, and from the range of $(79.8\%-86\%)$ to the range of $(79.5\%-90\%)$ for clients 5, 6, and 7. Note that the minimum coverage of client 7 in \verb!FedPerl! (at $0$ risk) is $30\%$, while it is $0$ coverage at $0$ risk for \verb!SSFL!. Client 6, on the other hand, achieves a minimum coverage of $56\%$ at $2\%$ risk. Utilizing our method achieves lower risk and better coverage in skin lesion classification.                            
\paragraph{\noindent\textbf{Reliability Diagram and Calibration Error}}
\label{secRD}
To investigate the uncertainty and models' calibration, we draw reliability diagrams and the expected and the maximum calibration errors in Fig.\ref{figRD}. We show the results for the federated models include ours. The numbers inside the sub-figures show the calibration error for each client. The numbers next to the model name show the averaged ECE and MCE errors for all clients. In each figure, we present the models' accuracy at different confidence intervals, such that the width of each bin represents the difference between the highest and lowest confidences. The figures show that our method improves the calibration for all models and reduces the errors significantly \cf Fig.\ref{figRD} (\verb!FedPerl! vs. \verb!FedAvg! models). 
While the most interesting and surprising results reveal that the lower federated model (Lower \verb!FedAvg!) is the most calibrated model after the upper model (Upper \verb!FedAvg!), such that it is better than \verb!SSFL! and \verb!FedPerl! respectively.  We can attribute this issue to the uncertainty of using unlabeled data during the training of both models (\verb!SSFL! and \verb!FedPerl!). In contrast to that, the lower and the upper \verb!FedAvg! models only trained on high-quality labeled data. Nonetheless, our model has better calibration errors than the \verb!SSFL!, where the ECE and MCE are 0.144 and 0.277 for \verb!FedPerl!, and 0.152 and 0.287 for \verb!SSFL!, respectively. Besides, \verb!FedPerl! outperforms the remaining baselines with considerable margins, further results are presented in Fig.\ref{figApp.RD} in \ref{app.rd}. Such lower calibration errors indicate more reliable and confident predictions for the \verb!FedPerl! over the other methods. 
Moreover, our experiments showed that applying peer learning produced a more calibrated model than \verb!SSFL!, \cf Fig.\ref{figRD} (\verb!FedPerl! vs. \verb!SSFL! models). Yet, after applying peer anonymization, a better calibration error is obtained, \cf Fig.\ref{figRD} (\verb!FedPerl(T=2)! vs. \verb!FedPerl(T=2) w/o PA! models). That implicitly means that the used peers are calibrated enough to produce more accurate pseudo labels than the ones generated from the clients individually.   
\paragraph{\noindent\textbf{Skin lesion qualitative results}}
\begin{figure}[!t]
\centering
\includegraphics[width=\textwidth]{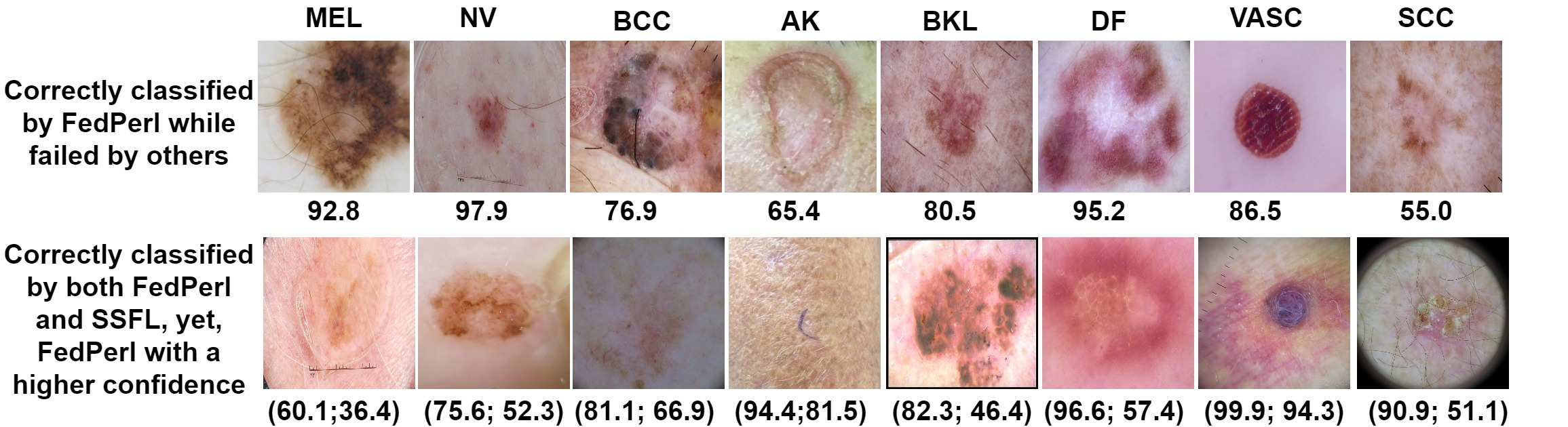}
\caption{Qualitative results. Sample predictions of FedPerl and SSFL for skin lesion. FedPerl confidence is shown below the images in the first raw, while the second raw shows the confidence for FedPerl and SSFL respectively.}
\label{figVS}
\end{figure}
Sample predictions of \verb!FedPerl! are shown in Fig.\ref{figVS}. The first row shows samples cases were classified correctly by \verb!FedPerl! but miss-classified by the other methods. Below each case, we show the prediction confidence. The first row shows the confidence for \verb!FedPerl!, while the second raw shows the confidence for both \verb!FedPerl! and \verb!SSFL! respectively. It is noticed that there are challenging cases, still, \verb!FedPerl! was able to classify them correctly, e.g. AK and SCC classes. The remaining cases were classified correctly with high confidence by \verb!FedPerl!, while they miss-classified by the others. The second row, on the other hand, shows cases were classified correctly by both \verb!FedPerl! and \verb!SSFL!, yet, \verb!FedPerl! achieves higher confidence. For instance, in BKL, DF, and SCC classes, the confidence margins are 35.9, 39.2, and 39.8, respectively.  
\subsection{Unlabeled Clients Scenario}
\label{secUClient}
\begin{table}[!t] 
\caption{The classification results for unlabeled clients scenario. +: with PA. $\dagger$:$\sim$\cite{yang2021federated}. $\ddagger$:$\sim$FedMatch\citep{jeong2020federated}}
\label{tableUL}
\centering
\resizebox{1\textwidth}{!}{
\begin{tabular}{l |l | c| c| c}
\hline
Setting & Model &F1-score & Precision & Recall\\
\hline
\hline
SSFL   &FedAvg$\dagger$&0.637(0.649)$\pm$0.121&0.647(0.649)$\pm$0.099&0.670(0.678)$\pm$0.120\\
         &FedMatch$\ddagger$&0.641(0.662)$\pm$0.131&0.653(0.657)$\pm$0.099&0.667(0.693)$\pm$0.134\\
w/o PA   & FedPerl(T=1)      &\textbf{0.644(0.662)$\pm$0.115}&\textbf{0.658(0.660)$\pm$0.078}&\textbf{0.674(0.688)$\pm$0.118}\\
w/o PA   & FedPerl(T=2)      &\textbf{0.644(0.670)$\pm$0.12}6&0.651(0.664)$\pm$0.100&\textbf{0.671(0.691)$\pm$0.130}\\
w/o PA   & FedPerl(T=3)      &\textbf{0.645(0.654)$\pm$0.117}&\textbf{0.655(0.657)$\pm$0.094}&\textbf{0.670(0.677)$\pm$0.123}\\
w/o PA   & FedPerl(T=4)      &\textbf{0.644(0.660)$\pm$0.124}&\textbf{0.655(0.665)$\pm$0.103}&0.668(0.678)$\pm$0.129\\
w/o PA   & FedPerl(T=5)      &\textbf{0.641(0.659)$\pm$0.129}&\textbf{0.655(0.660)$\pm$0.098}&0.668(0.681)$\pm$0.134\\
\hline
&FedMatch+          &\textbf{0.649(0.662)$\pm$0.118}&\textbf{0.655(0.659)$\pm$0.102}&\textbf{0.677(0.688)$\pm$0.121}\\
& FedPerl(T=2)      &\textbf{0.645(0.662)$\pm$0.119}&\textbf{0.654(0.659)$\pm$0.103}&\textbf{0.673(0.687)$\pm$0.119}\\
& FedPerl(T=3)      &\textbf{0.648(0.663)$\pm$0.118}&\textbf{0.660(0.669)$\pm$0.102}&\textbf{0.678(0.693)$\pm$0.120}\\
& FedPerl(T=4)      &\textbf{0.649(0.666)$\pm$0.124}&\textbf{0.656(0.663)$\pm$0.102}&\textbf{0.678(0.692)$\pm$0.125}\\
& FedPerl(T=5)      &\textbf{0.645(0.659)$\pm$0.114}&0.652(0.653)$\pm$0.096&\textbf{0.675(0.687)$\pm$0.118}\\
\hline
\end{tabular}
}
\end{table}

Till this experiment, we have trained our models to exploit the labeled and unlabeled data at each client. The previous setting is widely studied in the literature a.k.a the standard semi-supervised learning paradigm. In federated learning, however, a more challenging situation may appear to the surface in which the clients only have access to unlabeled data without knowing their annotations, see~\textit{Scenarios} in sec.~\ref{Scenarios} for more details. The results of applying this scenario to \verb!FedPerl! and our baselines are reported in Table \ref{tableUL}. Thanks to peer learning, our method enhances the performance of the baselines up to $0.8\%$ and $0.4\%$ compared to \verb!FedAvg! and \verb!FedMatch!, respectively. Moreover, an additional improvement of about $1.2\%$ is obtained after applying peer anonymization (see last four rows in Table \ref{tableUL}). That also holds for \verb!FedMatch! where \verb!FedMatch+! shows a relative improvement of about $0.8\%$ after applying PA. The better results are attributed to the aggregated knowledge from distributed similar clients who help the local models to overcome the missing of labeled data. 
\subsection{Generalization to Unseen Client Scenario}
\label{secGUC}
\begin{figure}[!t]
\centering
\includegraphics[width=\textwidth]{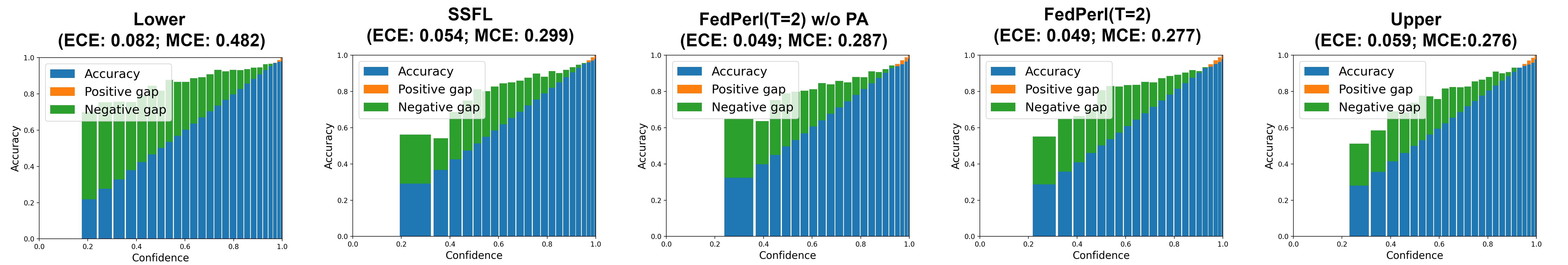}
\caption{Reliability diagrams and calibration errors on the ISIC20 dataset. FedPerl is more calibrated with lower calibration errors than the baselines.}
\label{figRD20}
\end{figure}
\begin{table}[!t] 
\caption{The unseen client scenario. The global models' classification results for FedPerl and the baselines on the ISIC20 dataset. +: with PA. $\dagger$:$\sim$\cite{yang2021federated}. $\ddagger$:$\sim$FedMatch\citep{jeong2020federated}}
\label{tableISIC20}
\centering
\resizebox{1\textwidth}{!}{
\begin{tabular}{l |l | c| c| c|| c| c| c}
\hline
\multicolumn{2}{c}{}&\multicolumn{3}{c}{Malignant}&\multicolumn{3}{c}{Benign}\\
\hline
Setting & Model &F1-score & Precision & Recall & F1-score & Precision & Recall\\
\hline
\hline
Lower    & FedAvg            &0.131&0.097&0.204&\underline{0.976}&0.985&\underline{0.966}\\
\hline
SSFL     & FedAvg$\dagger$   &0.161&0.114&0.274&0.974&0.987&0.962\\
    & FedMatch$\ddagger$   &0.160&0.113&0.278&0.972&0.987&0.954\\
w/o PA   & FedPerl(T=1)      &0.160&0.112&\textbf{0.279}&0.973&0.987&0.960\\
w/o PA   & FedPerl(T=2)      &\textbf{0.178}&\textbf{0.126}&\textbf{0.305}&0.974&0.987&0.962\\
w/o PA   & FedPerl(T=3)      &\textbf{0.166}&0.110&\textbf{0.339}&0.969&\textbf{0.988}&0.951\\
w/o PA   & FedPerl(T=4)      &\textbf{0.169}&\textbf{0.117}&\textbf{0.308}&0.972&0.987&0.958\\
w/o PA   & FedPerl(T=5)      &\textbf{0.166}&\textbf{0.120}&0.269&\textbf{0.975}&0.987&\textbf{0.965}\\
 & FedMatch+   &0.146&0.099&\textbf{0.281}&0.970&0.987&0.954\\
& FedPerl(T=2)  &\textbf{0.163}&0.113&\textbf{0.295}&0.973&0.987&0.959\\
& FedPerl(T=3)      &\textbf{0.167}&0.114&\textbf{0.308}&0.972&0.987&0.957\\
& FedPerl(T=4)      &\textbf{0.170}&\textbf{0.115}&\textbf{0.324}&0.971&0.987&0.956\\
& FedPerl(T=5)     &0.150&0.099&\textbf{0.305}&0.968&0.987&0.950\\
\hline
Upper   & FedAvg       &0.153&0.095&0.382&0.961&0.988&0.935\\
\hline 
\end{tabular}
}
\end{table}

The goal of this experiment is to investigate the generalization ability of the federated models to unseen clients. To achieve this, we collect the previously trained global models, including the baselines and \verb!FedPerl!, then we perform inference on the ISIC20 dataset. Note that this dataset consists of more than 33K images with two classes; malignant and benign. Take into consideration that the class distribution is highly imbalanced such that around 500 images contain malignant cases, while the remaining images have benign cases. Also, the models trained to distinguish between 8 classes making the direct inference a very challenging task. To resolve this issue, we perform two steps. First, we generate the eight-class predictions from the models. Then, we assemble these predictions into two groups. The malignant group contains melanoma, basal cell carcinoma, actinic keratosis, and squamous cell carcinoma classes. The benign group includes melanocytic nevus, benign keratosis, dermatofibroma, and vascular lesions. Then,  we generate our metrics as a binary classification task.

The results are reported in Table \ref{tableISIC20}. Interestingly, \verb!FedPerl! obtains the best malignant-class classification results outperforming the lower, the \verb!SSFL! including \verb!FedMatch!, and the upper bounds, with F1-score up to 0.178 for \verb!FedPerl! models. Note that, for clinical applications, the ability of a model to detect the true positive cases (malignant) is high relevant than detecting the true negative cases (benign) because the early detection of cancerous lesions reduces the treatment cost and the death rate. The ability of \verb!FedPerl! to classify the malignant and benign classes is also shown in the reliability diagrams and calibration errors, \cf Fig.\ref{figRD20}. We can see from the figure that \verb!FedPerl! is more calibrated and achieves better expected and maximum calibration errors than \verb!SSFL!. From these results, we show that \verb!FedPerl! has a better generalization ability to detect malignant cases than the baselines and \verb!FedMatch!. 
While we have seen in all previous experiments that applying PA to \verb!FedMatch! (denoted as \verb!FedMatch+!) always boosts its performance, this observation does not hold in this experiment. Specifically, the F1-score drops from $0.160$ to $0.146$. The same observation is found for some \verb!FedPerl! models.   
\subsection{\tariq{Comparison with SOTA in the Few Labeled Clients Scenario}}
\label{secFedIRM}
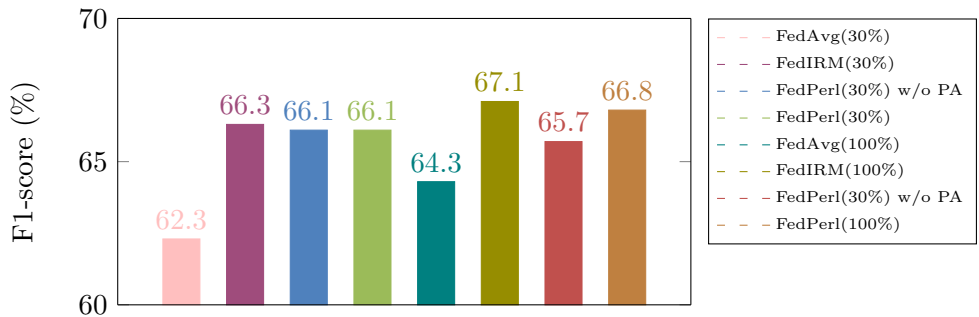
\begin{figure}[!h]
  \centering
  \begin{tikzpicture}
    \begin{axis}[
        width  = 0.5*\textwidth,
        height=1.5in,
        legend style={font=\tiny, solid},
        legend pos=outer north east,
        legend cell align={left},
        scale only axis,
        clip=false,
        separate axis lines,
        axis on top,
        xmin=0,
        xmax=9,
        xtick={1,2,3,4,5,6,7, 8},
        x tick style={draw=none},
        xticklabel=\empty,
        nodes near coords,
        ytick={60, 65, 70},
        ymin=60,
        ymax=70,
        ylabel={F1-score (\%)},
        every axis plot/.append style={
          ybar,
          bar width=14pt,
          bar shift=0pt,
          fill
        },
      ]
      \addplot[pink]coordinates {(1,62.3)};
      \addplot[ppurple]coordinates{(2,66.3)};
      \addplot[bblue]coordinates{(3,66.1)};
      \addplot[ggreen]coordinates{(4,66.1)};
      \addplot[teal]coordinates{(5,64.3)};
      \addplot[olive]coordinates{(6,67.1)};
      \addplot[rred]coordinates{(7,65.7)};
      \addplot[brown]coordinates{(8,66.8)};
    \legend{FedAvg(30\%), FedIRM(30\%), FedPerl(30\%) w/o PA, FedPerl(30\%), FedAvg(100\%), FedIRM(100\%), FedPerl(30\%) w/o PA, FedPerl(100\%)}
    \end{axis}
  \end{tikzpicture}
\caption{Comparison between our Method and FedIRM. While both methods achieve comparable results when the participation rate =30\%, ours show lower performance when 100\% of the participation rate. Still, FedPerl outperforms FedAvg.}
\label{figFedIRM}
\end{figure}

In this experiment, we conduct a comparison with \verb!FedIRM!~\citep{liu2021federated}; very recent work in \verb!SSFL! for the skin lesion classification. 
Notice that \verb!FedIRM! introduced a scenario where some clients are labeled while others are not. In addition, the training paradigm in \verb!FedIRM! assumed that all clients participate in the training in each round, \ie $PR=100\%$, which is not applicable in many cases. The vast majority of federated learning approaches assume that a random set of clients will participate in the training each round, which was our selection in this paper where the $PR=30\%$. Thus, to cover both cases, we present the results at  $PR=\{30\%, 100\%\}$. For our comparison, we opt \verb!FedAvg! and \verb!FedPerl(T=2)! models. Note that the hyperparameters are kept as in the previous experiments, while the results are reported in Fig.\ref{figFedIRM}. First, let us consider when $PR=30\%$. \verb!FedAvg! obtains F1-score equals 62.3 while \verb!FedIRM!, \verb!FedPerl! w/o PA, and \verb!FedPerl! achieve comparable results at 66.3, 66.1, and 66.1, respectively. Although our method outperforms \verb!FedAvg! ($PR=100\%$), we observe a slight relative drop in the performance, when we compare to \verb!FedIRM!, by 1.4\% and 0.3\% for \verb!FedPerl! w/o PA and \verb!FedPerl!, respectively.
That could be attributed to that \verb!FedIRM! only transfers the knowledge from labeled to unlabeled clients to guide the pseudo labeling process. However, this is not the case in our method where we utilize similar peers (regardless of their labels). Note that around 80\% of the clients are unlabeled in this particular scenario favoring the \verb!FedIRM! method. 
Still, \verb!FedPerl! outperforms \verb!FedAvg!, and extensive hyperparameters tuning could yield better performance for our method. For the same reasons, \verb!FedPerl! w/o PA, when  $PR=100\%$, achieves the lower results among \verb!FedPerl! models with F1-score equals 65.7, where more unlabeled clients were involved in the training. Yet, averaging the unlabeled peers might cancel their negative impact on the local model, as shown by \verb!FedPerl! with peer anonymization (PA) at F1-score = 68.7.
\subsection{Dynamic Learning Policy}
\label{secLP}
\begin{table*}[!t] 
\caption{Dynamic learning polices results. The classification results under two different settings. $C_{ISIC}$, $C_{HAM}$: the results at the community level. (mean F1-score). PA+/-: with/out Peers anonymization.}
\label{tablePol}
\centering
\resizebox{\textwidth}{!}{
\begin{tabular}{l|l|c c c c c |c|c c c|c|c c|c}
\hline
\hline
\multicolumn{15}{c}{The classification results when the clients contain labeled and unlabeled data (the standard SSL setting)}\\
\hline
\hline
Policy&Model/Client& 0& 1& 2& 3& 4&$C_{ISIC}$ & 5& 6& 7&$C_{HAM}$& 8& 9 & Avg\\
\hline
No Policy (baselines)    & PA- FedPerl(T=2) &0.735&0.731&0.725&0.737&0.739&0.733&0.805&0.850&0.839&0.831&0.582&0.729&0.747\\
&  PA+ FedPerl(T=2) &0.737&0.737&0.724&0.730&0.751&0.736&0.818&0.846&0.834&0.833&0.567&0.717&0.746\\
\hline
Validation Policy & PA- FedPerl(T=2)                      &0.729&0.727&0.724&\textbf{0.737}&0.746&0.732&0.814&0.845&0.819&0.826&0.571&\textbf{0.732}&0.744\\
 & PA+ FedPerl(T=2)                      &\textbf{0.743}&0.729&\textbf{0.736}&0.732&0.749&\textbf{0.738}&0.806&0.845&0.822&0.824&0.572&0.724&0.746\\
\hline
Gated Validation Policy & PA-(75) FedPerl(T=2)  &0.729&0.727&\textbf{0.738}&0.732&0.750&0.735&0.814&0.841&0.828&0.827&\textbf{0.598}&0.725&\textbf{0.748}\\
& PA+(75) FedPerl(T=2)                         &\textbf{0.746}&0.727&\textbf{0.737}&0.731&0.748&\textbf{0.738}&0.814&0.844&0.834&0.831&0.571&0.725&\textbf{0.748}\\
& PA-(85) FedPerl(T=2)                         &0.728&0.734&\textbf{0.740}&\textbf{0.738}&\textbf{0.760}&\textbf{0.740}&0.814&0.842&0.830&0.829&0.535&0.710&0.743\\
& PA+(85) FedPerl(T=2)                         &\textbf{0.738}&0.732&0.723&\textbf{0.742}&\textbf{0.755}&\textbf{0.738}&\textbf{0.818}&\textbf{0.850}&0.838&\textbf{0.835}&\textbf{0.596}&\textbf{0.729}&\textbf{0.752}\\
& PA-(95) FedPerl(T=2)                         &\textbf{0.739}&0.732&\textbf{0.735}&\textbf{0.737}&\textbf{0.754}&\textbf{0.739}&0.815&\textbf{0.850}&\textbf{0.839}&\textbf{0.834}&\textbf{0.583}&\textbf{0.730}&\textbf{0.751}\\
& PA+(95) FedPerl(T=2)                         &\textbf{0.747}&\textbf{0.745}&\textbf{0.731}&\textbf{0.738}&\textbf{0.752}&\textbf{0.743}&\textbf{0.818}&\textbf{0.851}&\textbf{0.839}&\textbf{0.836}&\textbf{0.596}&\textbf{0.731}&\textbf{0.755}\\
\hline
Gated Similarity Policy & PA-(75) FedPerl(T=2)  &0.734&0.721&\textbf{0.742}&\textbf{0.739}&\textbf{0.764}&\textbf{0.740}&\textbf{0.821}&0.843&0.832&0.832&\textbf{0.593}&0.714&\textbf{0.750}\\
& PA+(75) FedPerl(T=2)                         &\textbf{0.740}&0.725&\textbf{0.735}&\textbf{0.742}&\textbf{0.754}&\textbf{0.739}&0.811&0.841&0.827&0.826&0.580&0.699&0.745\\
& PA-(85) FedPerl(T=2)                         &0.727&\textbf{0.741}&\textbf{0.731}&\textbf{0.742}&\textbf{0.751}&\textbf{0.739}&0.812&0.846&0.825&0.828&\textbf{0.586}&0.716&\textbf{0.748}\\
& PA+(85) FedPerl(T=2)                         &\textbf{0.738}&0.735&\textbf{0.735}&\textbf{0.742}&\textbf{0.752}&\textbf{0.741}&\textbf{0.820}&\textbf{0.850}&\textbf{0.839}&\textbf{0.836}&\textbf{0.588}&\textbf{0.730}&\textbf{0.753}\\
& PA-(95) FedPerl(T=2)                         &0.734&0.728&\textbf{0.732}&\textbf{0.739}&\textbf{0.765}&\textbf{0.740}&\textbf{0.820}&0.845&\textbf{0.839}&\textbf{0.835}&\textbf{0.617}&\textbf{0.731}&\textbf{0.755}\\
& PA+(95) FedPerl(T=2)                         &\textbf{0.737}&\textbf{0.740}&\textbf{0.731}&\textbf{0.739}&\textbf{0.764}&\textbf{0.742}&\textbf{0.819}&\textbf{0.853}&\textbf{0.836}&\textbf{0.836}&\textbf{0.618}&\textbf{0.732}&\textbf{0.757}\\
\hline
\hline
\multicolumn{15}{c}{The classification results when the labeled data is only available on the server while the clients have no labeled data (the unlabeled clients or the disjoint setting)}\\
\hline
\hline
Policy&Model/Client& 0& 1& 2& 3& 4&$C_{ISIC}$ & 5& 6& 7&$C_{HAM}$& 8& 9 & Avg\\
\hline
No Policy (baseline) &PA+ FedPerl(T=2)                        &0.649&0.642&0.671&0.645&0.654&0.652&0.730&0.751&0.729&0.737&0.308&0.670&0.645\\
\hline
Validation Policy&  PA+ FedPerl(T=2)  &0.642&0.638&0.669&\textbf{0.647}&\textbf{0.678}&\textbf{0.655}&0.721&\textbf{0.752}&\textbf{0.740}&\textbf{0.738}&0.267&0.656&0.641\\
\hline
Gated Validation Policy & PA+(75) FedPerl(T=2)  &0.642&0.639&0.667&\textbf{0.647}&\textbf{0.659}&0.651&\textbf{0.740}&0.750&\textbf{0.745}&\textbf{0.745}&0.303&\textbf{0.678}&\textbf{0.647}\\
& PA+(85) FedPerl(T=2)                         &\textbf{0.669}&\textbf{0.657}&0.664&0.643&\textbf{0.666}&\textbf{0.660}&\textbf{0.740}&0.740&\textbf{0.744}&\textbf{0.740}&\textbf{0.340}&0.687&\textbf{0.655}\\
& PA+(95) FedPerl(T=2)                         &\textbf{0.653}&\textbf{0.650}&0.662&0.630&\textbf{0.669}&\textbf{0.653}&\textbf{0.743}&\textbf{0.751}&\textbf{0.733}&\textbf{0.742}&\textbf{0.343}&0.664&\textbf{0.650}\\
\hline
Gated Similarity Policy & PA+(75) FedPerl(T=2)  &\textbf{0.659}&\textbf{0.660}&\textbf{0.671}&\textbf{0.650}&\textbf{0.657}&\textbf{0.659}&\textbf{0.732}&0.749&\textbf{0.744}&\textbf{0.742}&\textbf{0.334}&\textbf{0.671}&\textbf{0.653}\\
& PA+(85) FedPerl(T=2)                         &\textbf{0.649}&\textbf{0.645}&\textbf{0.679}&0.636&\textbf{0.656}&\textbf{0.653}&0.728&\textbf{0.757}&\textbf{0.751}&\textbf{0.745}&\textbf{0.327}&\textbf{0.670}&\textbf{0.650}\\
& PA+(95) FedPerl(T=2)                         &\textbf{0.658}&\textbf{0.647}&\textbf{0.678}&\textbf{0.655}&\textbf{0.676}&\textbf{0.663}&\textbf{0.732}&\textbf{0.759}&\textbf{0.738}&\textbf{0.743}&\textbf{0.336}&\textbf{0.672}&\textbf{0.655}\\
\hline
\end{tabular}
}
\end{table*}

Previously, we have shown that the static peer learning policy is constantly beneficial to clients and communities. For instance, see the results in Table \ref{table2}. Also, we have shown that for the individual clients, who do not belong to any community, our method is still profitable, as for client 9. Yet, for other clients, \ie client 8, we have seen that peer learning, \verb!FedMatch!, and \verb!FedAvg! perform lower than the local model. Even though our model is better than the others. To resolve this issue, we propose, in section \ref{method:DLP}, a dynamic learning policy that controls the learning stream on the clients. The results are reported in Table \ref{tablePol}.
Due to the enormous amount of models that could be examined in this experiment, we opt for $\rho$ at $\{0.75, 0.85, 0.95\}$ and $T=2$. We generate the results for the standard semi-supervised and unlabeled clients scenarios. Our baseline in this experiment is our model \verb!FedPerl(T=2)! as our goal is to compare with the static policy, and we do not see any need to include the previous models which already compared with \verb!FedPerl(T=2)!.
\subsubsection{The standard semi-supervised learning results}  
\paragraph{\noindent\textbf{Validation Policy}}
First, by comparing overall results, denoted as $Avg$ in Table \ref{tablePol}, we notice no significant improvement in the performance for both models; PA($\pm$) \verb!FedPerl(T=2)!. On the other hand, lower results are obtained for the $C_{HAM}$ community. For instance, the F1-score dropped from $0.831$ and $0.833$ to $0.826$ and $0.824$, respectively. In contrast, a comparable result at $0.732$ or a slight enhancement at $0.738$ are obtained for $C_{ISIC}$. Besides, the clients' results are inconsistent regardless if they belong to a community or not. While we notice boosting for the clients 0, 2, 3, and 9, the remain clients have lower results. Further, we notice no positive influence on the results when applying PA. 
\paragraph{\noindent\textbf{Gated Validation Policy}}
While there is not much benefit in the previous policy, the results in this experiment show a consistent improvement as the gateway threshold $\rho$ increases. For instance, the overall results boosted up to $0.2\%$, $0.6\%$, and $0.9\%$ when $\rho=0.75$, $0.85$, and $0.95$, respectively. 
The consistent improvement also found at the community level when $\rho$ is larger than $0.75$, with better results at $\rho=0.95$. While an increase reaches $1\%$ is noticed for $C_{ISIC}$ clients starting from $\rho=0.75$ with PA model \ie PA+(75) \verb!FedPerl(T=2)!, the increase is seen starting from PA+(85) model for $C_{HAM}$ clients with F1-score reaches $0.836$. 
In general, the clients' results get boosted by our gated validation policy. In the beginning, when $\rho=0.75$, clients 0, 2, and 8, show better performance comparing to the baseline. Then, more clients are included when $\rho=0.85$ until all clients show improvement with our model PA+(95) with F1-score at $0.752$ at client 4. These results confirm the same behavior found in communities' results. A more discussion on the individual clients' results, \ie 8 \& 9, reveals that the combination of PA with values of $\rho=\{0.85, 0.95\}$ achieves more reliable F1-scores. Even though our model PA-(75) obtains the highest score for client 8, the results for other clients are not of the same quality. 
In summary, we present in this experiment that our gated validation policy improves the overall, communities', and clients' results demonstrating its advantage. More importantly, the results of client 8 were boosted from $0.567$ to $0.596$ at PA+(95) model.                  
\paragraph{\noindent\textbf{Gated Similarity Policy}}
This policy is different from the earlier one in using the similarity between the client and its peers as a gateway to control peers' participation instead of using the global validation dataset. We notice that the general behavior is similar to the preceding one. Though, better results are obtained at different levels, especially for client 8, whose reported F1-scores are equal to $0.617$ and $0.618$ on models PA$\pm$(95), which are better than the former ones by $1.9\%$ and $4.7\%$ respectively. The similarity in the results is justified because both policies proposed to manage the learning stream on the clients, especially the individual ones, which has been shown in both strategies. A gated similarity policy brings more stability to all clients and better accuracy for client 8.                    
\subsubsection{The unlabeled clients' results}
We have shown in the past section that the validation policy has no potential improvement, while the combination of the gated methods with PA usually obtains the best performance. Therefore, and for simplicity, we opt to report only the results with the PA technique. 

After analyzing the second part of Table \ref{tablePol}. We notice that the results of validation policy are improved by F1-score equals to $0.655$ and $0.738$, for $C_{ISIC}$ and $C_{HAM}$ respectively. Yet, the overall results decreased by $0.4\%$. The individual results, on the other hand, vary between the clients. While clients $3, 4, 6,$ and $7$ show an enhancements, clients' $0, 1, 2, 5, 8,$ and $9$ accuracies are decreased.   
In contrast to the previous results, we observe a constant improvement of gated policies in the overall accuracy from $0.647$ to $0.655$ for the validation with PA+(75) to similarity with PA+(95) gated models, respectively. Note that all models from both policies accomplish better results than the baseline model. The communities' results, on the other hand, show comparable results, yet better than the baseline, for both strategies with some advantages for the similarity models. While the individual improvement is distributed among the clients in the gated validation approach except for client 2, it is intelligible in similarity models, especially in PA+(95) model. Moreover, both individual clients; 8 \& 9, show steady improvements in all similarity models, yet, client 9 suffers from lower performance in gate validation with $\rho$ larger than $0.75$. However, the maximum gain appears for client 8 in the PA+(95) gated validation policy with F1-score equals $0.343$.

\section{Discussion}
Our method; \verb!FedPerl! compiles many concepts such as semi-supervised learning, federated learning, peer learning, committee machine, and learning policies to devise a novel framework for skin lesion classification tasks. We show through extensive experiments and evaluation metrics that our method has superior performance over the baselines in the standard semi-supervised labeled and unlabeled clients settings.
\paragraph{\noindent\textbf{FedPerl simplicity \& performance.}}
A key feature of our method is simplicity. Implementing and applying our method is direct and can be implemented with a few lines of code. The computational cost to calculate the similarity between the clients is negligible, thanks to our strategy which computes the similarity on extracted features rather than on the whole weight parameters. Such that for a model with $l$ layers and $\omega$ weights parameters, where $\omega \ggg l$, the cost of our similarity is O($l$) $\lll$ O($\omega$), note that $\omega$ could be million of parameters. From another perspective, the experiments show that \verb!FedPerl! is more calibrated and outperforms the baselines including the \verb!SSFL!, thanks to the peer learning we propose, where \verb!FedPerl! exploits other clients by interacting with their experiences. As a core component in our method, peer anonymization reduces the communication cost while enhances performance. Additionally, \tariq{it improves the clients' privacy by hiding their identities}. Yet, a non-avoidable cost is still property in peer learning.
\paragraph{\noindent\textbf{Similarity.}}
Clients' communities are shaped implicitly based on the similarities between the clients. To measure the similarities, we exploited models parameters to profile the clients. Yet another approach to quantify the similarity is to use a server-side validation set as it has been utilized in \verb!FedMatch!. While we have shown through different experiments that our method of finding the similarities outperformed the one that depends on validation set~\ie \verb!FedMatch!, another drawback is that the availability of validation datasets at the server side is a challenging task. Further, we have shown the importance of peer learning and our similarity in the random peers experiment. Still, the representational similarity is an open research direction in federated learning.  
\paragraph{\noindent\textbf{Orthogonality.}}
Another main property of the PA technique is that it can be implemented directly to other methods, which are similar to ours, with negligible effort. We have shown through different experiments that applying PA to \verb!FedMatch! is resulted in a better model, \ie \verb!FedMatch+!. While the new model achieves better accuracy, it also reduces the communication cost comparing to the original one. %
\paragraph{\noindent\textbf{Privacy.}}
Our anonymized peer is designed by aggregating/averaging the model parameters of the top $T$ similar peers. This process generates a virtual model that is not related to a specific client and offers a harder target for attackers seeking information about individual training instances~\citep{mcmahan2017communication,orekondy2018gradient}. Nevertheless, a privacy guarantee for aggregated models (not individuals) is an open issue and has not been thoroughly investigated in the community and mathematical analysis is yet to be proven. 
{
\paragraph{\noindent\textbf{Local Updates.}} 
While the local models' weights are continually updated during the training, the peers' ones remain intact. A natural question could be if such a procedure might poison the models, especially with larger iteration updates? While such concern is of high importance, we have designed our method to alleviate this problem by training the local model and keeping the peer models intact to avoid any poisoning. Also, we employed an MSE loss as a consistency-regularization, \verb!FedVC! approach in the federated learning, and our dynamic policy, especially if the local model is quite different from the peers and has been trained for more local iterations.}
\paragraph{\noindent\textbf{FedPerl communities \& committee size.}}
Figure \ref{figcommunities} shows that \verb!FedPerl! clusters the clients into communities based on their similarities. The overall performance for each community gets boosted by \verb!FedPerl!, \cf Tables \ref{tableCommu} and \ref{table2}, which is attributed to the knowledge sharing. We have noticed that the community performance is related to the committee size \ie $T$. While changing $T$ has an insignificant effect on $C_{ISIC}$ performance, its effect is clear on $C_{HAM}$. Thus, a natural question would be, what is the ideal committee size? Our experiments show that as long as $T$ below the actual community size, the overall performance is rather stable, \cf $C_{ISIC}$ in Table \ref{tableCommu}. Once $T$ exceeds the community size, the performance starts decreasing, \cf $C_{HAM}$ in Table \ref{tableCommu}. We associate this with the probability of including external peers as we increase $T$, which might have a negative influence on the local models/sites of the community, see sec.\ref{exp:BC}. While the cluster/community size can be defined by the cardinality of the spectral clustering of the similarity matrix, yet in more practical scenarios, setting $T$ to a value larger than the community size is impractical. The trade-off between the committee size and the performance needs further investigation. 
\paragraph{\noindent\textbf{\tariq{FedPerl clustering.}}}
\textcolor{black}{
The clustering in the literature means grouping similar data and assigning labels to them. Because we use this word frequently in our paper and to resolve any ambiguity, we provide the following interpretation. First, we do not use any clustering method nor it is defined heuristically or fixed at the beginning of the federated learning. Hence, we do not assign labels to the clusters, but rather we want to highlight that our similarity matrix works effectively to force similar peers to learn from each other. At the beginning of the training, the clusters, \ie learning from similar peers, are dynamically changed, which is explained by the small numbers in each row in Fig.~\ref{figcommunities}, where the darker colors or smaller numbers values represent lower frequencies. However, as the training proceeds, these clusters are evolved to force the similar peers to learn from each other more frequently, which is shown by the brighter colors or higher numbers values in the same figure.} 
\paragraph{\noindent\textbf{FedPerl \& individual clients.}}
The clustering produces individual clients who do not belong to a specific community \ie clients 8 \& 9, which confirms the reality. The effect of \verb!FedPerl! is diversified between those two clients. While client 9 makes use of \verb!FedPerl!, a drastic drop in the performance of client 8 was noticed, which could be attributed to the class distribution mismatch. This indicates that \verb!FedPerl! may not fit non-iid scenarios. 
Yet, combining \verb!FedPerl! with works that are handling the distribution mismatch (non-iid) problem would be a promising direction of research \citep{li2018federated,zhao2018federated,li2021fedbn,zhang2021federated}. 
On the other side, one nice property has been shown by our experiments that \verb!FedPerl! is less sensitive to the noisy clients than the standard \verb!SSFL! and \verb!FedAvg! methods (\cf Table.~\ref{table2}), which could be attributed to the learning schema of selecting similar peers in \verb!FedPerl!. \tariq{In our experiments, we found out that inductive bias coming from similar in-distribution clients did not hurt the global model, it rather improved the global model performance. Having said that, Out-of-Distribution (OOD) client, e.g., client 8, has shown to harm the model's performance. If there is a strong inductive bias from a couple of OOD clients, this potentially might hurt the global model. One might need to consider a smarter way of aggregation for such OOD clients. However, this is out of the scope of this manuscript.} 
\paragraph{\noindent\textbf{FedPerl \& unlabeled clients (Scenario \#2).}}
In a more challenging experiment, which is unique in federated learning, we trained our models utilizing labeled global data and unlabeled local ones. \verb!FedPerl! also shows the best results comparing to the baselines thanks to our peer learning strategy, which enforces additional knowledge to the clients besides the global one exploited via federated learning. Further, applying PA produced more stable results and higher accuracy. 
\paragraph{\noindent\textbf{FedPerl \& unseen clients (Scenario \#3).}}
In another part of our experiments, we have tested our method on the ISIC20 dataset. The low performance for all models can be attributed to two things; i) Class Mismatch: the models were trained on 8 classes while ISIC20 contains only two classes with severe class imbalance (500 malignant vs. 32.5K benign), and ii) Domain Shift: none of the models proposed to address the domain shift problem between ISIC19 and ISIC20. In this experiment, we tried to show how SSFL models perform in such a challenging situation. The results showed that our model is still better than all baseline models in skin cancer classification shedding the light on the generalization capability. This is attributed to the \verb!FedPerl! is learning more powerful and discriminative representations for the minority class by aggregating the peers' knowledge and experiences. While we attributed the better performance of our model than \verb!FedMatch! to the similarity matrix that we utilized such that our method picks more accurate peers to the local model than the \verb!FedMatch! approach. In the current version, \verb!FedPerl! does not have a specific property that handles the class imbalance. 
{
\paragraph{\noindent\textbf{Learning from few labeled clients (Scenario \#4).}}
The comparison with a SOTA method reveals that our method is on par with \verb!FedIRM! when part of the unlabeled clients participate in the training. Yet, that is not the case when all clients are involved, which is a rare setup. We attributed that to the quality of the pseudo labels generated with the help of unlabeled peers to ones generated with the help of labeled clients. Nevertheless, combining both approaches in a joint or a co-training setup could be an interesting research direction and might lead to better performance.}              
\paragraph{\noindent\textbf{Learning Policy.}}
Our first strategy depends on a static peer learning policy that involves best $T$ peers based on their similarities. While this policy is effective in the communities and clients, it suffers in performance when countered by an ODD client. To resolve this issue, we proposed, in this paper, more dynamic and adaptive policies. Specifically, the successful policies employed a gateway to control the learning rate from peers. The participation is measured based on either a global dataset or how similar the peer is to the client. Only the clients who pass a predefined threshold can participate in the training. We found that the results of the two policies are somehow similar with advantages to the one based on similarity. Yet, and most importantly, the performance of the OOD client gets boosted from both policies.  Because we do not have control or can not anticipate the in-distribution from out-of-distribution clients, the selection between the static and dynamic methods goes toward the dynamic ones. Even if we know the clients, the results show that the dynamic policy betters the static one. On the other hand, our preference between the validation or similarity gated policies goes toward the similarity. In most cases, the global validation data is not available, which prevents us from applying the validation policy. Further, the gated similarity policy produces more consistent and stable results. Though, the trade-off between the global and local benefits could be the decision-maker in real-life scenarios. \tariq{Our dynamic learning policies are considered heuristic ones, however, they were proposed to address a problem that we noticed in our previous work~\cite{bdair2021fedperl}, where the performance of some individual clients has not improved by the federated learning. We could achieve that by utilizing the global validation dataset or the client similarities. However, to provide a comprehensive study addressing any potential questions from the reader, we tested three different policies. Note that the three policies are separate and work independently. Besides, they have shown to be effective (\cf~Table \ref{tablePol}).}

\section{Conclusion}
\label{conclusion}
In this paper, we propose a semi-supervised federated learning framework for skin lesion classification in dermoscopic images. Our method; \verb!FedPerl! overcomes the limitations of the previous works, inspired by peer learning from educational science and ensemble averaging from committee machines. We show a real-life application of our method that fits the complexity of the medical data \ie data heterogeneity, severe class imbalance, and an abundant amount of unlabeled data. Our database consists of 71,000 skin lesion images obtained from 5 public datasets. The testing environment consists of the standard semi-supervised setting and a more challenging and less investigated scenario where clients have access just to the unlabeled data. \tariq{Our method is on par with the state-of-the-art method in skin classification in the standard federated learning, \ie random set of clients participating in the training} and outperforms the baselines and other \verb!SSFL! by 15.8\%, and 1.8\%, respectively. Moreover, \verb!FedPerl! demonstrates less sensitivity to noisy clients and has better generalization ability to unseen data.  Besides, we propose the peer anonymization (PA) technique. PA is a simple and efficient approach to create an anonymized peer and hide clients' identities. PA enhances performance while reduces communication costs. We show that our method is orthogonal and easy to implement to other methods without additional complexity. In this paper, we investigate two learning policies; a fixed policy that selects the top similar peers, and a dynamic and more adaptive one that controls the learning stream on the clients. We have shown that both strategies are effective with advantages to the dynamic one. Thus far, we exploited the model parameters as similarity measurement, while we could employ different techniques to profile the clients. Further, we could investigate the privacy guarantee for aggregated models as future work.


\acks{T.B. is financially supported by the German Academic Exchange Service (DAAD).}

%
\ethics{The work follows appropriate ethical standards in conducting research and writing the manuscript. This work presents computational models trained with publicly available data, for which no ethical approval was required.}

\coi{The authors declare no conflicts of interest.}

\bibliography{refs}

{\noindent \em Remainder omitted in this sample. }

\end{document}